\documentclass[12pt]{article} 
\usepackage{nips13submit_e,times}
\usepackage{url}
\usepackage{graphicx}
\usepackage{xcolor}
\usepackage{microtype}
\usepackage{anyfontsize}
\usepackage[hidelinks]{hyperref}
\usepackage{amssymb}
\usepackage{wrapfig}

\title { \textbf{\fontsize{25}{27}\selectfont Modeling the Mind: A brief review}}

\author{\vspace{15mm}
\\
{\vspace{5mm} \fontsize{15}{18} \selectfont Gabriel Makdah} \vspace{5mm} \\ 
{\fontsize{15}{18}\texttt{\color{gray!90!black} {\fontfamily{cmss}\selectfont {\textls{gabriel.makdah@etu.univ-lyon1.fr}}}}} \vspace{5mm} \\
\\
Universit\'e Claude Bernard Lyon 1, France\vspace{35mm}\\ 
\LARGE\setlength{\fboxsep}{15pt} 
\setlength{\fboxrule}{1pt}
\fbox{$I$ - Computational Foundations} \vspace{4mm} \\ 
}

\nipsfinalcopy 

\begin{document}

\maketitle
\thispagestyle{empty}
\newpage

\setcounter{tocdepth}{2}
\tableofcontents
{\setlength\parindent{1cm}
\setlength\parskip{4mm}

\newpage
\begin{center}
\fontsize{13}{15}\textbf{Preface}

\textit{Modeling the Mind: a brief review} is an annual review, available for free on arXiv.org, whose aim is to help students and researchers unfamiliar with the field of neuroscience and computational neuroscience gain insight into the fundamentals of this domain of study. Creating an accurate simulation of the mind is no easy task, and while it took brilliant minds decades to advance us to where we're at right now, we are still ways off our final goal. It is therefore imperative to have more research carried out in this multidisciplinary field, taking in help from researchers in biology, neuroscience, computer science, but also mathematics, physics, chemistry and imaging, in order to speed up this process and tip the scales in our favor for the upcoming decades. This annual review hopes to provide the required information for anyone who is considering this domain as his future endeavor. 

The reviews will be tackling relatively global characteristics at first in order to familiarize the reader with the basic foundations, and will be getting progressively more specific and in tune with current research in the upcoming parts. This is Part I. It will contain basic information about the computational aspect of this field, and will attempt to explain why certain concepts are generally agreed upon, and the intuition behind them, going through the essential founding works.
\end{center}
\newpage

{\begin{center} 
\textbf{Abstract}
\end{center}}

\noindent The brain is a powerful tool used to achieve amazing feats. There have been several significant advances in neuroscience and artificial brain research in the past two decades. This article is a review of such advances, ranging from the concepts of connectionism, to neural network architectures and high-dimensional representations. There have also been advances in biologically inspired cognitive architectures, of which we will cite a few. We will be positioning relatively specific models in much broader perspectives, while also comparing and contrasting their advantages and weaknesses. The projects presented are targeted to model the brain at different levels, utilizing different methodologies.

\section{Introduction}

Computational neuroscience has now existed for several decades. It started off as an attempt to model individual neurons' firing mechanics, with the Hodgkin-Huxley model[30] in 1952, as well as neurons' learning models, with Hebbian learning[24] in 1949. Today, scientists use computer simulations not only to model some of the brain's traits, but also to predicate some of its behavior, and study new and emerging aspects of it. The results produced using simulations often lead to testable predictions in the real world.\\
	\indent Neural networks were created by researchers who wanted to model architectures that were similar to that of the brain. From constructing nodes, to learning algorithms such as backpropagation, the brain has greatly influenced neural network development. Ever since their conception, neural networks exhibited features that were remarkable for computers. To this day, we are still surprised by their expanding catalog of capabilities. They are used in tasks such as speech recognition, image recognition, learning grammars and languages, robotics, and a wide array of subjects, that require a human-like thought process to conduct. All of this makes artificial neural networks the prime contenders to modeling the mind.

\noindent In the upcoming sections, we will discuss some of the theories behind neural networks, their foundations, the most widely used algorithms, as well as some theories about their application to the field of neuroscience.
Section 3 will be dedicated to examining computer patterns, and representations in order to take maximal advantage of the computable space. The brain, and its whole specters of variety will be presented in section 4 in order to have a better understanding of it, before progressing any further. The concepts of computationalism, and connectionism, two theories for modeling the brain, will be discussed in section 5. In section 6, we will present artificial neural networks, and talk about the most widely used architectures and algorithms, as well as the advantages of each one of them. Section 7 will go over vector symbolic architectures and some basic operations and models to optimize storage. Section 8 will serve as an overview to the brain's working memory, as well as the modeling of the pre-frontal cortex's memory system, a key component to working memory. We will be talking about the main problems that artificial brain architectures should be tackling, as well as  give examples on how research has been aiming to solve them in section 9. The final section, section 10, will serve as a conclusion to the article. We will discuss what an optimal model of the brain would look like following the concepts that were discussed in this review.\\
	\indent But first, let us give an overview of what made digital computers possible, and talk briefly about some of the basic principles behind them in section 2. 
\vspace{.5mm}

\section{The foundations of computers}

\subsection{Hardware}

The grounds of today's digital computers lie in the Von Neumann architecture.\\
	\indent The Von Neumann[84] architecture is a computer model described 60 years ago, whose success has made computers a ubiquitous part of our lives. The architecture depicts an electronic digital computer with different parts. The first, and arguably the most important of these, is the processing unit, which contains an Arithmetic Logic Unit, ALU for short, as well as processor registers. The control unit houses instruction registers and a program counter. It helps directs operations and instructions to the processor. Memory is used to store data and instructions, either in rapidly accessible Random-Access Memory, RAM for short, or slower, but more capacitive external mass storage, such as hard disk drives, and more recently, solid state drives. The machine also needs input hardware in order to take queries from the operator, equipment such as keyboards,  mice, or pressure sensors are used. In order to display the results of its computation, the machine expresses itself through output hardware, such as a monitor, speakers, or a robotic arm.\\
	\indent Data, and the instructions that manipulate data, are entities of the same kind. They are stored in the same set of addresses and data buses. The Von Neumann architecture has only one memory space. 

\noindent A lot of attempts have been made to program computers for the kind of flexible intelligence that characterizes human behavior, however with no remarkable success. This has lead many to wonder whether a different computing architecture is needed, and whether the problem was stemming from a lack of software optimization, or an inherently limiting bottleneck in the Von Neumann architecture. Several new architectures were presented throughout the years, each tackling different issues. Some are found widely in use today, which prove that the Von Neumann model is not the be all and end all of architectures.\\
	\indent An example of such models would be the Modified Harvard architecture. This model has 2 separate memory spaces, compared to only one in the Von Neumann architecture. The first memory space is used to read data from, and write data to the memory. The second set of addresses and data buses is used to fetch instructions. This results in an increase of throughput, seeing as the memory is no longer being limited by memory spaces. While the system is executing one instruction, it could be fetching the next one at the same time.

\noindent In our case, the burden falls on representation, the kinds of entities that the computer uses in order to process its computations. If our goal was to model the mind, then we ought to be using similar representations to those used by the brain. Instead of using low-dimensional binary vectors as conventional computers would do, the basic idea here is to compute with large vector patterns, called high dimensional random vectors. We will be tackling this subject next, but first, let us give a general definition of computing.

\subsection{Computing}
Computing is the transformation of patterns by algorithms which are governed by rules. Patterns, also known as representations, are the configuration of ON and OFF switches in a sequence. In other words, they are the sequencing of bits. An algorithm helps compute a new pattern based on the previous strings of patterns. Patterns can be meaningful when they correspond to either abstract entities, or things in the world, such as numbers, words, images.\\
	\indent To transform patterns, computers have circuits, such as the adder circuit for summing two numbers, the OR circuit, the AND circuit...etc, whose interpretations are the same as their uses in ordinary language. The materials that computers and circuits are made of are subordinate. That is to say they are not a major contributor to making the logic possible. Materials can be changed as cost of production and development varies, leading to better constituents. The logical design can be separated from the materials and the hardware. This intelligent design can be seen not only in man-made computers, but in humans themselves. Although we are all capable of comparable computing power, our underlying architectures are very different. No two people have the same neural hardware, yet the logical design still holds from one person to the other.
	
\section{Representations}
Computers use patterns formed by strings of 0s and 1s. These are called binary vectors, and are used in binary representations. The goal of a certain representation is to be able to discriminate between any two objects. In other words, the bit patterns representing two different values must differ from one another.\\
\indent The meanings which patterns hold are not apparent in their standalone binary vectors. Instead, it's the relations between different patterns that determines their possible use for computing. This is also true for the brain. A standalone neuron being activated does not yield much, if any, work done. It's the complete structure of inter-relational patterns that generate computation.\\
	\indent Choosing the correct representation is subject to trade-offs and compromises. There are no perfect representations that would be suitable for all tasks. The base-2, binary, system is extremely efficient in arithmetic operations and is therefore used in computers. Our brains very probably use a different kind of representation, with different trade-offs, better suited for vital functions, compromising the speed and precision of arithmetic tasks.\\
	\indent However, brain-like abilities and functions can still be demonstrated in computer \mbox{simulations} using nothing but binary patterns and operations. This proves that, up to a certain point, the precise nature of the individual units don't matter as much as the properties deriving from them through the specified operations. 

\noindent Representations can be done through a variety of ways, depending on the type of tasks, and the computer architecture that is at hand. Scalars are one-dimensional vectors. They are represented using higher dimensional vectors in computers: the scalar 46260 is represented using a 16bit integer, which is a 16-dimensional binary vector. In the next sections, we will talk briefly about the basic features underlying the brain's representation models, and then generalize to what our model should incorporate.

\subsection{Dimensionality}
A huge number of neurons and synapses form the brain's circuitry. Each neuron is connected to tens of thousands of other neurons. There are hundreds of thousands of neurons that fire each second. The amount of information that is being exchanged inside our brains, between neurons, and between regions, is extremely large. To match such high content and rate of exchange, we will shy away from scalars and low-dimensional vectors. The information in the brain is encoded on larger elements than these. This leads us to explore the possibility of high-dimensional vector use, vectors that are thousands of bits long[37].\\
	\indent A high-dimensional vector is represented in a high-dimensional space, which groups all the vectors of one representation together. Just like a 2-dimensional vector is represented in a 2 dimensional space, an N-dimensional vector is represented in an N- dimensional space. High dimensional modeling started off several decades back under the forms of artificial neural networks, parallel distributed processing and connectionism. We will be exploring all these areas later in this review. First however, let us expand a bit on these high-dimensional representations.
	
\subsection{High-dimensional representations}

The space of representations make up the sets of units with which a computer computes. This space is low-dimensional in conventional computers, where the memory is usually addressed in 8-bit bytes, and the ALU applies its arithmetic operations on 32-bit vectors.
	The high-dimensional representational vectors and space aren't necessarily binary. They can be ternary, real, or complex. They can also be further modulated using probabilistic distributions, limited ranges of values, and sparse modeling. Sparse modeling corresponds to loosely coupled systems. Many components are either zero, or negligibly small. This saves a large amounts of memory, and speeds up data processing, while maintaining the advantages of high-dimensionality[35]. Before advancing any further, let us take an example of a low-dimensional vector.
}

{\fontsize{12}{18} \texttt{\color{gray!90!black} {\fontfamily{cmss}\selectfont \textls {Let us consider a 3 bit string:
	$$(1,0,1)$$
\indent This string is also considered to be a $3-dimensional$ vector.\\
\indent It can be represented in a $3-dimensional$ space.
\\
\\
\indent Such a three-dimensional space can contain $2^3 = 8$ patterns.}}}}

\noindent To showcase the difference in orders of magnitude between a low, and high-dimensional representation, let us now take an example of a high-dimensional binary vector.

{\fontsize{12}{18} \texttt{\color{gray!90!black} {\fontfamily{cmss}\selectfont \textls {Let us consider a 20,000 bit string:
	$$(1,0,0,1,0,1,1,0,0,1...etc)$$
\indent This string is also considered to be a $20,000-dimensional$ vector.\\
\indent It can be represented in a $20,000-dimensional$ space.
\\
\indent A binary $20,000-dimensional$ space consists of:
\begin{center}
 $2^{20,000} = 3.98 * 10^{6020}$ such patterns.
\end{center}}}}}
{\setlength\parindent{1cm}
\setlength\parskip{4mm}   
\noindent Because of the discrete nature of vector representations, all vectors are equally spaced. Distances can be measured between points using the Euclidean metric, or the Hamming distance. The Hamming distance is the number of places at which two binary vectors differ.  Between two 20,000-dimensional vectors, the Hamming distance cannot exceed 20,000 bits. This can be normalized to 1, in Hamming distance, by dividing it by the dimensionality of the vector. Because of the nature of the representational space, the large majority of distances between two vectors are concentrated at 0.5 (or 10,000 bits in our example). In statistical terms, picking two random vectors, there is over 99,99\% chance that the distance separating them is between 0.47 and 0.53. If we take a third vector, it will also differ from the first two by around 10,000 bits. These vectors are considered as unrelated, and are said to have a dot product between them of mean zero.\\
	\indent Because of these characteristics, if we were to take a large number of vectors, such as 10$^9$ vectors that are 20,000-dimensional, as an example, this will still be a very small percentage of the total number of vectors in vector space. Noisy vectors would still be identifiable among the large number of chosen vectors, as long as the chosen vectors are all randomly spaced between each others. Related, or similar, concepts can be coded in vectors that have smaller Hamming distances separating them. This would make it so if the noise was too great, and we were unable to recuperate the original vector, we would at the very least obtain a closely matching, or similar, one. 

\noindent A cognitive system such as the brain includes several representational spaces, of different dimensions each, which are used for different tasks. As we will see later in this review, some neurons exhibit the characteristics of low-dimensional vectors, while others demonstrate properties of high-dimensional representations.

\subsection{High-dimensional memory}

The memory stores the data, and set of instructions necessary to run, and manipulate a program. A classical memory architecture is made up of memory locations, which are an array of addressable registers, each of which hold a string of bits. The contents of the memory locations is made available for use by probing the memory using the location's address, which is also a string of bits. In our case, a memory that would store 20,000 bit vectors would have to be addressed by 20,000 bit vectors, and would have a total of 220,000 memory locations in order to store vector space as a whole.\\
	\indent This overflow of memory locations would make it possible to retrieve a particular set of data using it's address, or even an approximation of its address through noise and error. There are two storage modes, each with distinct properties and advantages.
	
\subsubsection{Heteroassociative and Autoassociative memory}

Heteroassociative memory is based on the mechanism of storing a memory $\mu$ using the pattern of $\alpha$ as an address. $\mu$ can thereafter be retrieved using either $\alpha$, or a noisy approximation of $\alpha$, such as $\alpha'$, $\alpha''$...etc. This type of memory makes it possible to store sequences within memories and their addresses. By making $\alpha$ the address for the memory $\mu$, and $\mu$ the address for the memory $M$, we can sequentially recall memories $\mu$ and $M$ using only $\alpha$ as an initial address.\\
\indent First coined as Correlation Matrix memories[39], this type of architecture was not able to correctly perform a recall task using noisy addresses. Hetero-associative memories saw a big leap with Bidirectional Associative Memories[40]. In this model, subtle noise in memory addresses could be overcome, and memory recall worked with forwards, and backwards passes, by storing the memory in vector pairs.

\noindent The other type of memory storage is autoassociative memory. Autoassociative memory is achieved by storing a memory $\mu$ using $\mu$ itself as the address. Just as previously described, $\mu$ can be recalled by $\mu$, or by a noisy version of $\mu$, such as $\mu'$, $\mu''$...etc. This type of addressing therefore works as a noise filtering technique. It is used in several content-addressable memory models, such as the Hopfield network[32], which guarantee a convergence to a pure, non-noisy, memory, through several iterations.\\
\indent Going from $\mu'''$, to a less noisy $\mu''$, to $\mu'$, we can then finally obtain the stored memory $\mu$. However, depending on the address used, its proximity with the initial value, and how noisy a particular vector is, convergence to a false pattern could still occur.

\section{The brain and diversity}

There are two big types of neurons that are present in the brain, each of which has their own representations. The first type of neurons perform extremely simple and specific tasks, such as color recognition or space orientation. These neurons behave as bits, or low-dimensional vectors. Their firing is straight forward and predictable using what we know today as basis for their computation algorithms.\\
   \indent The other type of neurons is mixed selectivity neurons. These neurons are most notably present in the thinking, planning and higher order regions of the brain. Unlike classical selectivity neurons, these don't respond exclusively to one stimulus or task. Instead, they react in different ways when confronted with a wide variety of stimuli[3]. They are essential for complex cognitive tasks and give us a computational boost and a cognitive flexibility. Up until very recently, it was thought that these neurons fired in random, stochastic waves. Their behavior did not conform to what we knew about neurons.

\subsection{Mixed selectivity}

Mixed selectivity neurons are different from classical selectivity neurons. They are nonlinear, highly heterogeneous, seemingly disordered, and difficult to interpret. However various teams are trying to reverse engineer their structures, by collecting data through various neurophysiological experiments. In order to observe these features, single neuron recordings must be performed in situ. Because of the invasive nature of the procedure, experiments have been restricted only to behaving animals, particularly to the prefrontal cortical region, a region known for its implications in thought, behavior and decision making.\\
\indent The results seem to indicate that the populations of mixed selectivity neurons encode distributed information. This feature was not observed in the classical neurons, which are confined to single tasks. 
	
\noindent The neural representations are also high-dimensional, where the dimensionality represents the firing rate of the neurons. This makes it possible for these neurons to have a large set of input to output relations. Through these features, neurons can generate a rich set of dynamics and task solving capabilities[66]. The nonlinear mixed selectivity is sufficiently diverse across neurons that the information about task types could be extracted from the covariance between neural responses. Even when the classical individual neurons were experimentally devoid of any useful information, the knowledge and task solving capabilities of the primate were conserved. In order to remove classical selectivity, noise was added to every classical neuron, equalizing their average responses, and nullifying their effects. After removing this classical selectivity, a large drop in accuracy was observed in the early stages of the trial. However the accuracy increased later on as the trial progressed, and more information was gathered[67].

\noindent It is interesting to emphasize that these same properties were observable using high-dimensional representations, simulated on a classical Von Neumann machine using binary vectors. It's essential to get a close look into the brain's circuitry and representations through neuroscience. But it is equally important to then learn to abstract away from the neurotransmitters, membrane potentials, spike trains, and other physical attributes of the brain, and instead tackle the resulting behaviors. This will help us understand the underlying mathematical principles based on them. In other words, the logical design must be separated from neural realization.

\subsection{Robustness}

It is widely known that humans lose gray matter as they age. In fact, it's been observed that there's a physiological 10\% to 30\% reduction of the cortical gray matter density, in normal individuals, between ages 7 and 60[78]. While some individuals may lose as much as 1/3rd of their brain, this goes on with little to no perceivable change. With all these variations, and the significant day-to-day loss of neurons, our every day life goes by uninterrupted. The neural architecture is therefore very tolerant to component failure. That is, unlike classical binary representations in computers, where even a single bit of difference could either result in system errors, or lead to significantly different results.\\
	\indent Redundant representations are a hallmark of high-dimensional vectors. The higher the dimensionality, the higher will the proportion of allowable errors be. As discussed earlier, patterns can differ by a certain number of units, yet they could still be considered equivalent if the representational-space was large enough.
	
	\noindent For maximal robustness, and an efficient use of redundancy, the information should be distributed equally among all units. This would lead to a steady degradation of the stored information. This degradation is proportional to the number of failing bits, irrespective of their types or positions. While we would like to employ a robust architecture, it is also essential to take into account the plasticity that the brain exhibits. Modeling is therefore done in a holistic manner, which is a perfect trade off between robustness and plasticity, optimizing neural resources and redundancy[85].

\subsection{Randomness} 

Brains are highly structured organs, but many details are left to randomness. Network formation is not very constrained, with a strong component of randomness reigning on it ever since its genesis. The distance between nodes, and their connections follow an independent random distribution, making no two brains structurally identical[71]. Any two brains are therefore incompatible at the level of hardware and internal patterns.\\
	\indent Not only is there a structural randomness, but also constant spontaneous neuronal quantal release. This is due to the random probability associated with transmitter release[91]. Random potentials are therefore stochastically created, and are known as synaptic noise. Synaptic noise can aid, or impair, various features such as signal detection, neural performance, and may shape certain patterns.
	
	\noindent For a correct neural simulation of the brain, the system should have randomness ingrained in it. The model should be built using stochastic patterns and connections, by drawing vectors in a pseudo-random fashion. The actual patterns themselves aren't of any significance. What does matter for computation and consciousness are the relations between patterns within the system. Patterns for two similar concepts should be similar in one system, whereas two similar, or identical, concepts should theoretically have different patterns in different systems.
	
\section{Computationalism, and Connectionism}

Research on intelligent systems has split into two different systems, the classical symbolic artificial intelligence paradigm, also known as computationalism, and the connectionist AI paradigm.

\subsection{Computationalism}

Symbolic AI is at the forefront of computationalism. It maintains that the correct level at which we should model intelligent systems, such as behaving animals, or the human mind, is through symbols. A symbol is an entity that can have arbitrary designations, such as objects, events, and relations between objects, events, or both. Each symbol corresponds to one, and only one, entity at any given time.[55] Symbols may be combined together to form symbol structures, such as expressions or sequences. In a conventional computer, a symbol is represented by a variable, which can be instantiated, by making it refer to a specific entity, or uninstantiated, where it doesn't refer to any specific entity, and can therefore be rebound.

\subsection{Connectionism}

Connectionism is split into two parts: localist representations, and distributed representations. Each of them has their respective advantages and short-comings. We will be detailing them in the upcoming section.

\subsubsection{Localist representations}

In localist representations, a single unit, or neuron, represents a single concept. It is relatively close to the idea of symbolic AI, except instead of binding objects to units, we are binding features of objects to units. This gives much more freedom in representing entities with similar properties, but comes at the cost of harder variable binding, and a longer processing time.\\
	\indent Variable binding can be achieved in several ways, most notably using signature propagation[88]. A separate node is allocated for each variable associated with each entity, and a value is allocated to represent a particular object, it is the signature of the object it expresses. The signature can then be propagated from one node to the others in feed-forward, or backward chaining processes depending on the task at hand. Signatures will also need to be manipulated during state transitions, certain actions or orders, and variable rebinding. This can be done through vector addition, or multiplication.\\
	\indent Another way to bind variables in localist representations is through phase synchronization, which utilizes the temporal aspects of activations. An activation cycle is divided into multiple phases. Each node fires at a particular phase, and can be either an object node, which holds a certain value, or an argument node, which holds a variable. Each phase represents a different object involved in reasoning. Objects have their own assigned phases, and fire in a consistent manner. When an argument node fires at the same phase as an object node, the variable is bound to its value. After that dynamic binding is accomplished,  which forces the variable node to fire synchronously in phase with the constant node it is binding[43]. Reasoning is therefore a propagation of temporal patterns of activation. 

\subsubsection{Distributed representations}      

In distributed representations, each entity is represented by a pattern of activity that is distributed over many units. Each unit may also be used to represent many different entities[75]. \\
	\indent One way to implement distributed representations is through the use of microfeatures. Microfeatures[26] are a more detailed approach to the items of the domain. Unlike the first order features used in localist representation, these are composing elements of the features described above. Different entities can be represented through a distinctive pattern over these elements, making it possible for two different features, or two different objects, to share some microfeatures.\\
	\indent Another way to realize distributed representations, that doesn't involve the use of semantic microfeatures, is coarse coding[69]. Coarse coding imposes two conditions on the features that are to be assigned to units. The features must be coarse, that is, they must be general enough to be shared by more than one entity, and have an extended definition. The assignments of the units must overlap and superpose, in such a way that one unit is involved in representing several different objects. This redundancy leads to a highly fault-tolerant network. The coarse code is a robust representation; the failure of one, or even several components will not make the system unusable. These redundancies also help the system provide a higher than normal degree of accuracy. Similarities between objects will therefore be reflected by similarities among their representations.\\
	\indent As is obvious,  systems here work at sub-conceptual levels, and entities don't have a precise, complete, or formal description. Instead, features are represented by a pattern of activity over a number of different units, scattered in the network in a non-organized fashion. Because of this, the system has a complex internal structure that is crucial for its correct operation. This may be the most optimal way for the system to go through processing, however it will be very hard for a human operator to decipher and interpret these structures. As a result, variable binding has proved to be exceptionally difficult. A lot of attempts have been made, of which we will cite a few now, while other more successful attempts will be explained in more detail, later in this review.

\noindent The Distributed Connectionist Production System[79], DCPS for short, manipulates symbolic expressions using conditional rules. Rules act as rails, and specify which expressions should be moved in, or out, of the working memory of the DCPS. The distributed representation is done by means of coarse coding. However, the main limiting factor is the presence of only one variable in the whole system. Variable binding is therefore very limited, and can only be performed on a special set of terms. By using a memory that is large enough, we may avoid immediate limitations by temporarily storing our variables there, at the cost of computational time. However this is a non optimal solution that does not encompass all use case scenarios. It comes at the cost of a severe bottleneck originating from the system, trading computational time for simplicity, with the lack of variables provided.\\ 
	\indent Tensor product representations[77] have also been used to implement variable binding. Vectors are combined using vector addition and aperiodic convolutions. This creates a more complex object out of 2 or more basic features. Each object is therefore made up of a set of variables, which are the arguments or roles, and each role has a value, or filler. A vector can represent either a variable, or a value. Variable binding is accomplished by forming the tensor product using aperiodic convolution. In order to represent a whole entity or item, tensors can be added together. Pattern matching can be carried out using these tensor products, but full variable to value unification cannot be achieved through this method. However, this concept is the origin of much of the modern view on distributed representations, as we will be seeing later.
	
	\noindent In order to achieve an efficient level of binding variables to their values in connectionist representations, rapid variable binding is essential. This implies creating variables on the fly, and binding them to their correct values in a virtually instantaneous manner, before deploying them, and using them according to their context. However, that comes with a wide range of problem of its own[21].

\subsection{The necessity of variable binding}

As discussed, in order to meet the requirements of synaptic plasticity and speed of the brain, rapid variable creation and binding should be performed in distributed representations. There have been numerous attempts that were made in order to solve this problem, each with its own trade-offs.\\
\indent Eliminative connectionism was introduced in order to contour the problem of rapid variable creation, without tackling it head on. The objective of this approach was to completely avoid using localized or explicit variables. However, in order to match the brain's capacity to induce and represent identity relations, as well as concatenation functions, variables are needed for the manipulation of data. Explicit variables are also needed to achieve a decent level of generalization, and functional mapping[44]. The brain has a remarkable ability when it comes to representing relations between variables that are novel in our experience with variables that are familiar to us. Not only that, but humans are able to induce a general pattern from a novel test case in a remarkably fast time. Let us demonstrate this through an example:
}

	{\fontsize{12}{18} \texttt{\color{gray!90!black} {\fontfamily{cmss}\selectfont \textls {\indent \indent Consider the following set of inputs and outputs:\\
\indent  Input: A car is a // \hspace{20.2mm} Output: Car \\
\indent  Input: A house is a // \hspace{14.7mm} Output: House \\
\indent  Input: A capacitor is a // \hspace{7.2mm} Output: Capacitor \\
\\
	\indent \indent \indent  What would the Output be for:\\
\indent  Input: A juitrekfluit is a // }}}}

\noindent The solution to this test case is apparent. In less than a second, a Human subject can figure out the answer to this novel case, a word which he had never seen before. One may argue that the semantic meaning associated with these words might bias the human operator, making it unfair to compare him with a machine that doesn't have any capacity at understanding phrases or words. We will therefore present another more involved, and random, example:

	{\fontsize{12}{18} \texttt{\color{gray!90!black} {\fontfamily{cmss}\selectfont \textls {\indent \indent Consider the following set of inputs and outputs:\\
\indent Input: jutrika setek bolur hiuna // \hspace{5.2mm}	Output: Bolur \\
\indent Input: jutrika setek nyata hiuna // \hspace{4.7mm}	Output: Nyata \\
\indent Input: jutrika setek jinski hiuna // \hspace{5mm}	Output: Jinski \\
\\	
	\indent \indent \indent What would the Output be for:\\
\indent Input: jutrika setek kouyte hiuna // }}}}
		
{\setlength\parindent{1cm}
\setlength\parskip{4mm}
\noindent This generalization capacity, induced in hundredths of a second when faced with even completely novel structures, does not arise from training a network the moment the sample data was presented. There is simply not enough time to train a neural network this quickly, especially at the level of complexity exhibited by the brain. If synaptic weights were indeed being changed, this can only be done through Long Term Potentiation, or Short Term Potentiation. Long Term Potentiation requires dendritic spines growth, and therefore takes hours to develop. Short Term Potentiation is a transitory alteration of the synaptic vicinity, and takes around 10 seconds to develop. If what was observed here was not due to new changes in synaptic weight, yet was novel enough to surely not have a neural network specifically trained to suit it, then it very probably relies on some pre-existing skill, acquired through the training of subnetworks within the brain at a past date. These networks are being used with their pre-existing variables, to which novel values are being bound rapidly, on the fly.\\
	\indent A practical solution to this problem may reside in VSAs, or Vector Symbolic Architectures, which permit effective variable binding, and can lead to generalization in the presence of suitable algorithms[54]. This group of architectures assumes that each value, and each variable is represented by a distributed activation pattern within a vector. An additional vector can be used to encode the result of binding a value, to its variable. However even in such cases, it is impossible to predict how a network would assign and operate through novel tasks on its own, as weight vectors gained through regular training may not suffice for generalization to new novel cases[21]. We will be detailing VSAs later in this review.

\section{Artificial Neural Networks}

Artificial neural networks, ANN for short, are an adaptive computational method that learn a particular task through a series of examples. They can be used to model various cases of nonlinear data. Just so, it happens to be that most physical phenomena are inherently nonlinear. This makes artificial neural networks an excellent candidate to model real world occurrences. The test samples are multivariate data represented by several features and components, that is to say, they are represented by high-dimensional vectors[83].\\
	\indent ANNs were once deemed a curse brought onto computer science. They produced astonishingly accurate results, but using high-dimensional vectors came at the cost of an exponentially growing number of function evaluations. This occurrence was due to the lack of optimization of older algorithms, which used exhaustive enumeration strategies[6] that were sub-optimal for such tasks. Currently, we are better equipped to deal with high-dimensional neural networks through dynamic programming[33], conditional distribution of features among hidden layers[8], and generally better performing and more optimized neural network structures such as recurrent neural networks and restricted Boltzmann machines[1].\\
	\indent Old ANN algorithms were developed mostly for low-dimensional computing. Nowadays, with the era of massive automatic data collection systematically helping both researchers and programmers obtaining a large number of measurements, new interest has been kindled in researching high-dimensional modeling[13]. This motivation was also supported by parallel discoveries in neuroscience, which helped generate computer models based off of the efficient computational abilities of our brains. The first of such models were PDPs.

\subsection{Parallel Distributed Representations}

Parallel distributed processing[46], PDP for short, is a type of artificial neural network that took its inspiration from the processing model of the brain. In situ, neural connections are distributed in parallel arrays, in addition to serial pathways. Functions are not computed sequentially in series, but are rather processed concurrently and parallelly to each others. More than a single neuron is sending signals at a time, and up until the mid-1980s, this was a rarely used feature in the digital computer.\\
	\indent As was discussed earlier, in the Von Neumann architecture, each processor is able to carry out only one instruction at a time. In PDPs however, information is represented in a distributive manner over different units, and more than one process can occur at any given time. Memory is not stored explicitly, but rather is expressed in the connections between units, similar to how neural synapses work. Learning can occur through gradual changes in connection strengths through experience. This system also exhibits all the advantages of distributed representations.
	
\subsection{Restricted Boltzmann Machines}

Restricted Boltzmann machines, RBM for short, are a type of artificial neural networks that specialize in deep-learning[76]. An RBM has only 2 node layers: A visible layer, that receives input, and a hidden layer, which allows the occurrence of probabilistic states.\\
	\indent An RBM uses a generative probabilistic distribution over a set of inputs, in order to extract relevant features. More than one RBMs can be stacked on top of each others, in order to gradually aggregate groups of features. Each RBM within the stack deals with features that are progressively more complex. Through these numerous layers, several groups of features are conceived, which help decode the presented concept into smaller models of representations. The more stacks there are, the better they handle complex tasks. This, however, comes at the cost of computational time. A network composed of aggregated RBM stacks is called a deep-belief networks, and is considered to be one of the more prominent breakthroughs in artificial intelligence research and deep learning.\\
	\indent Each layer communicates with both the previous and the subsequent layer. Nodes do not communicate laterally with other nodes of the same layer. While each hidden layer serves as an effective hidden layer for the higher nodes, it also acts as a visible, input layer for the lower nodes. An RBM is fed one test trial at a time, without any prior knowledge of its nature. Learning can be supervised, where the system infers functions and features from labeled data, or unsupervised, where the network attempts to find a structure on its own, among unlabeled data. The connection weights between two nodes that lead to a correct conclusion become stronger, while the weights that lead to wrong answers become weaker.
	
\noindent	However, restricted Boltzmann machines are not deterministic. Given a particular set of units, initial conditions, inputs, and connections, the network won't always arrive to the same conclusion, due to its stochastic nature. The network operates through probabilistic distributions over patterns, by doing several passes over the original test set. States are updated accordingly, with a high probability of the network to converge, and a low probability for it to diverge[70]. This is unlike deterministic networks, such as Hopfield Networks, where the network is programmed to always, and consistently, attempt to converge through each step. What this leads to is a higher operation time, at the cost of lower chance of getting stuck at local minimums. That is, there is a higher chance to find the "absolute truth" in stochastic models, because their randomness is enough to, almost, force the network to settle at the lowest state possible.\\
	\indent The most widely adopted algorithm for use in restricted Boltzmann machines is Markov Chains[12], which are logical circuits, that connect two or more states using guided and learned probabilities. Probabilities exiting one state add up to 1. This algorithm is sequential, and given one state, it can dissect what the next state would be, and predict subsequent states and sequences.

\subsection{Recurrent Neural Networks}

\subsubsection{General Overview}

Recurrent neural networks, RNN for short, are a powerful class of artificial neural networks where connections between units form a directed cycle. The initial inspiration for this model came from neurophysiological observations. Neurons were not only feedforward, they were also interconnected. Signals between neurons carried over by all or none principles. Thus was born the concept of RNNs. Nodes are not only linked using forward chaining connections, there are also lateral connections, as well as connections from current layers into past layers. This binary stream[47] had a remarkable short-term memory for its time, and had influences that stretched far beyond neural networks. It even influenced the development of the digital computer by Von Neumann. \\
\indent The basic architecture of recurrent neural networks is the fully recurrent neural network. Each node has directed connections to every other node. There are many models and algorithms that seek to better the computation and learning capabilities of RNNs. Apart from their use in the computational and mathematical fields, RNNs have experienced a large adoption rate in brain modeling. They are not only a way for researchers to reproduce the mind's capabilities digitally, but also a way for them to theorize features which have not been experimentally observed yet through the traditional means of neuroscience or psychology.

\subsubsection{Turing Completeness}

RNNs' inherent representational power made them an attractive candidate for universality, otherwise known as Turing Completeness. Turing Completeness can be achieved by proving that the system at hand can compute each and every function computable by a Turing machine. In other words, this means the system should be able to calculate any recursive function, use any recursive language, and solve any problem solvable using any effective method of computation or algorithm[81][82].\\
	\indent At first, computational universality was achieved using recurrent neural networks with a finite number of neurons, and high order connections, by combining their activations through multiplicative means[63]. However, this is an unattractive candidate for practical use because of the leaps of data that must be calculated. Such a model would consume a large amount of memory space and its convergence time would grow exponentially with the complexity of a given problem. Such a model is sub-optimal on many different levels. Computational universality was also achieved in linear RNNs using additive connections, assuming an unbounded, and infinite number of neurons[90]. This is also not practical, as having an infinite set of nodes is physically not possible.
	
\noindent In order to assume Turing completeness, one can simply prove that the system can simulate a Turing machine. In the case of finite, low-order, recurrent neural networks, this can be done by proving that there exists a network, started from an inactive initial state, that can lead to the correct output from an input sequence, using a recursively computable function. The input could also be removed entirely, to be then encoded into the initial state of the network. This should lead to a valid output, and correct results. Computational universality was therefore proved to stand true in low order connection, finite neural networks[74]. An estimated total of 886 neurons (870 processor nodes, and 16 input and output nodes) is needed to reach this universality.

\subsection{Using an external stack memory}

A recurrent neural network with no hidden layers is capable of learning state machines, and is at least as powerful as any multilayer perceptron[19]. It contains an inherent internal memory that's relatively small. This internal memory, stored in weights between the nodes is not enough to store all the details that would be necessary for intermediate or long-term use. In order to boost memory storage and long term performance, one could theoretically use an internal memory stack. However, this would require a large amount of resources, and would come in the form of additional nodes and connections, which would in turn exponentially increase computational time[64]. An optimal memory stack would therefore be external. It would allow for easy modulation, as well as continuous communication with the network, which permits the use of classical and continuous optimization methods, such as gradient descent.\\
	\indent An example of this is the Neural Network push-down automaton (NNPDA, [11]). This architecture consists of a set of fully recurrent neurons representing the network's state, and permitting classification and training. State neurons receive their knowledge from three distinct sources. First off, they receive their information from input neurons, which register input fed to the network by sources external to the system. Secondly, from read neurons, that interact with the external memory stack, and keep track of the symbols that are on top of the stack. And thirdly, they receive feedback from their own recurrent connections, seeing as this is a recurrent neural network. The network is connected to the stack through a non-linear error function, which approximates results that hold with high probability. 

\noindent Although it is possible to learn simple tasks using first-order networks, higher order nets are a necessity for learning more complex features. As we've grown accustomed to when it comes to complexifying computational algorithms, this comes at the expense of a longer processing time, and slowed convergence. First-order networks are by no means complete, they lack several features required to correctly order stored memory, as well as isolators, to help fend off any unwanted interference. We will see how this problem could be surpassed using LSTMs, and NTMs.

\subsection{Long Short-Term Memory}

As we stated before, recurrent networks can use their feedback connections in order to store representations of recent input events in the form of activations. This is a very useful feature when dealing with tasks requiring multiple inquiries over a short-term memory.\\
\indent With conventional algorithms such as back-propagation through time[86], the error signal is scaled proportionally to the number of steps it is propagated back. This leads the error to blow up, growing exponentially and leading to oscillating weights, bifurcations and unstable learning when the error is positive[60]. If the error is negative, this in turn leads to vanishing weights, which induces excessive learning times when bridging events through long time lags, rendering the modification of existing weights a time consuming challenge to be dealt with[7]. The conventional backpropagation method is simply too sensitive to recent distractions.

\subsubsection{The basic architecture}
	
To avoid the problems that come with back-propagation, long short-term memory, LSTM for short, was introduced[29]. It's a second-order recurrent neural network that enforces constant error flow, by propagating it through dedicated constant error carrousels(CEC). LSTM uses a linear constant activation unit to store the error, as opposed to the non-linear sigmoid activation functions used for the other computing units.\\
	\indent Because the constant unit follows a different scaling path than the rest of the units in the network, units connecting to it will be receiving conflicting weight updates. This makes efficient learning difficult to achieve, and unstable to maintain. To counteract this phenomena, we add multiplicative input and output gates. The input gates protect the memory content stored in the constant unit from outside perturbation; the error is therefore only edited when it needs to be updated. Output gates protect the rest of the units in the network from the irrelevant information stored inside the constant error carrousel; the error therefore only affects the network when it is recalled and needed. The gates are context sensitive nodes that can be trained, using gradient descent, to decide when to overwrite or access the values in the CEC.\\ 
\indent	A "memory cell" is built around the CEC, using a central linear unit, with a fixed self-connection, and delimited by the input and output gates discussed earlier. Several memory cells can share the same gates, and form a memory cell block, which is a more efficient architecture that facilitates storage. Error inside the memory cell blocks do not get scaled when propagated back further in time because of their linear constant activation functions. This eliminates exploding or disappearing error values, and stabilizes long-term memory.

\noindent This architecture can handle noise, distributed representations, and continuous values, while increasing the ability to generalize over untrained values better than older implementations of neural networks, including BPTT[86] and RTRL (real-time recurrent learning[68]).

\begin{figure}[t]
\begin{center}
\includegraphics[scale=0.7]{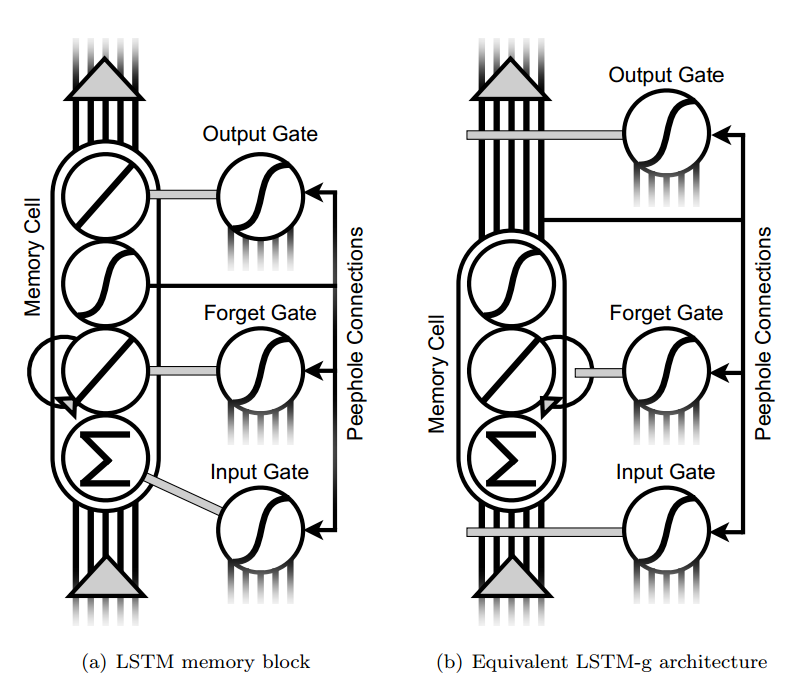}
\caption {{\fontsize{10}{8}\selectfont  {\textbf {Comparison between an LSTM memory cell block, and its equivalent in LSTM-g.} Black lines: Weighted connections. Grey lines: Gated connections. In the LSTM model(a), the input is first summed, then gated as a whole by the input gate. The output is also computed as a whole, all subsequent units receive the same modulated output. Peephole connections project from an internal stage in the memory cell to the controlling gate units, only the final value of the memory cell is visible to other units. In the LSTM-g model, the inputs to the memory cell are gated individually before being summed. The output leaves the memory cell unmodulated, and is then individually modulated by the output gate to other cells. Peephole connections connect to the recurrent units of the memory cell, and are able to see its intermediate state. This allows for greater flexibility. Image courtesy of [51].}}}
\label{fig:lstm1}
\end{center}
\end{figure}

\subsubsection{Extensions to the basic LSTM model}

LSTM, being one of the more promising RNN architectures of its time, received several improvements along the years. Forget gates[17] were added along side the input and output gates. Forget gates modulate the amount of activation a memory cell keeps from its previous time-step. They are able to force a quick memory dump in memory cells that have completed their tasks. This helps in rapidly keeping units effective, and augments the proportion of efficiently used and recruited nodes during later computations.

\noindent Another improvement to LSTM which quickly became a staple in computation are peephole connections. Peephole connections[18] are links between the three types of gates, and the memory cells. In the basic LSTM model, when the output gates are closed, the memory cell's visible activity is null. The information contained inside the CEC is therefore hidden from its associated gates, and surrounding nodes, making it impossible to efficiently control the information flow into, and out of the memory cell block. These peephole connections enable the cell's unmodulated state to be visible to outside units. This improves the network's performance significantly by reducing random instances of gating and probabilistic editing, which used to happen due to insufficient information.\\
	\indent In order to expand LSTM beyond second-order RNNs, Generalized LSTM \mbox{(LSTM-g, [51])} was introduced. In this architecture, all the units in the network have an identical set of operating instructions. A unit relies only on its local environment to determine what role it will be fulfilling. Units can be input gates, output gates, forget gates, memory cells, or any combination of these. Every unit is also trained using the same algorithms. That is in contrast with conventional LSTM, where each type of unit had its specific set of algorithms. A number of other changes were also made in order to improve flexibility. This helps improve performance, and broadens the applicability of the LSTM networks in machine learning. LSTM-g can be shown compared to the regular LSTM architecture in Figure \ref{fig:lstm1}.
\vfill

\subsection{Long Term memory}

Recurrent neural networks stand out from other machine learning methods due to their ability to learn and carry-out complicated transformations of data over extended periods of time. But one aspect that was largely neglected in computer science and research was the use of logical flow control and external memory. As we've discussed, RNNs are extremely effective when it comes to storing recent memories. With the advent of LSTM, intermediate-term memory storage became a lot more efficient. However, this did not significantly improve networks' storage capacities. During long, and complicated tasks, networks should have a way to store large quantities of information, while efficiently maintaining control over their flow.

\subsubsection{Neural Turing Machines}

Neural Turing machines[20], NTM for short, is a neural network architecture that attempts to deal with the problems showcased earlier. NTMs are enriched with a large addressable memory, that can be trained using gradient descent, to yield a practical mechanism for machine learning. For maximum effectiveness, the network will have to operate on rapidly-created variables, which are data that are quickly bound to memory slots. Their architecture is illustrated in Figure \ref{fig:2}.\\
\indent	Neural Turing Machines interact with the external world using input and output vectors, just like all neural networks. However they also interact with a memory matrix, using an attentional process to selectively read and write operations. Every component of the architecture is differentiable, and instead of addressing a single chosen element per cycle, the network could learn to interact with a variable number of elements in the external memory bank, as it sees fit. This is done through blurry read and write operations, determined by an attentional focus mechanism. This constrains them, and makes it possible to attend sharply to the memory at a single location, or weakly to the memory at many locations.\\
\indent	Read vectors are stored in the memory matrix's column, while the number of rows is equal to the total number of memory locations. Writing is done in two parts. First using an Erase Vector, a full memory location can be entirely, or partially emptied to make room for subsequent data. After that, the network uses an Add Vector to write to the memory. Both Erase and Add vectors are differentiable, computing their current state from their previous state, weight vectors and input orders. The presence of two separate vectors, both being independently subject to learning, allows a fine-grained control over which elements in each memory location are modified and how. 

\begin{figure}[t]
\begin{center}
\includegraphics[scale=0.7]{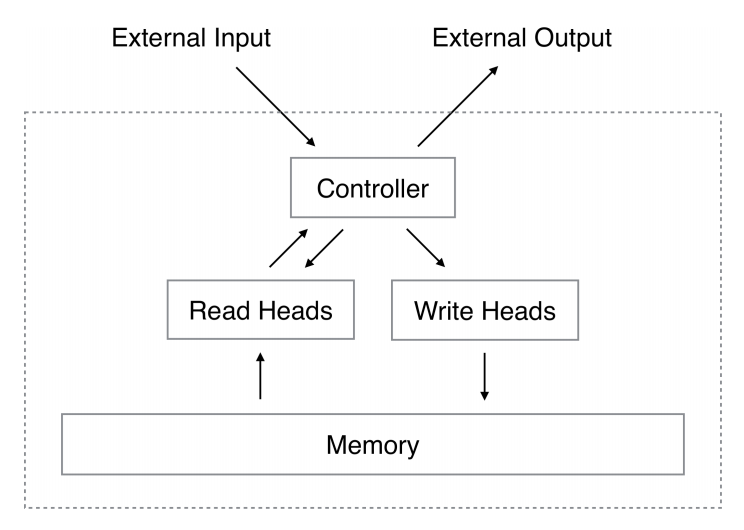}
\caption {{\fontsize{10}{8}\selectfont  {\textbf {Neural Turing Machine architecture.} The controller, or neural network, receives the input from an external environment, and emits an output in response. The controller is also able to read, and write from a memory matrix, using read, and write heads. Image courtesy of [20].}}}
\label{fig:2}
\end{center}
\end{figure}

\noindent The weightings arise from the combination of content-based addressing and location-based addressing. Content-based addressing focuses attention based on the similarity between the calculated, noisy values, and the noiseless values present in the memory space. This makes it possible to use approximations to speed up calculations, leaving it to the autoassociative memory to retrieve the exact value and clean up the noise[32]. Location-based addressing is used when values with arbitrary content are employed. Since it is not possible to predict these values, yet they still need a recognizable name and address, the values are addressed by location, and not by content. This is designed to facilitate both simple iterations across the locations, and random-access jumps[31]. The content of a memory location could also include information about the location of another value inside it.  This allows the focus to jump, for example, to a location next to an address accessed by content.\\
\indent	Either a feedforward, or recurrent network can be used here, each with its own advantages. While the RNN allows the controller to mix information across several time-steps of operation and permits a larger number of read and write heads, a feedforward controller confers greater transparency of the network’s operation, because its patterns are easier to interpret by a human operator. This transparency comes at the cost of an imposed bottleneck due to the reduced number of read and write heads.	

\noindent This network architecture leaves several free parameters for the operator to decide upon, such as the size of the memory matrix, the range of location shifts in the location-based addressing, the number of read and write heads, as well as the type of the controller network. All of these parameters serve to make the Neural Turing Machines[20] as adaptable as possible for a number of different tasks.
\vfill

\subsubsection{Memory Networks}

Another solution to the lack of long-term memory in neural networks are Memory Networks[87]. This architecture was conceived around the same time as the Neural Turing Machines described earlier, by a different, and independent research team, indicating a recent surge in interest in the use of external memories as an extension to neural networks. The model operates in similar ways to the NTM, by combining machine learning inference with an external memory component that can be read and written to.
	This is done through four learned components, them being an input feature map, an output feature map, a response element, that converts the output into the desired format, and a generalization component, which updates old memories given the new input and the current state of the network.

\section{Vector Symbolic Architectures}

Vector symbolic architectures, VSA for short, are high-dimensional vector representations of objects, relations and sequences, used in machine learning. A good representation helps systems compute orders of magnitude faster, making possible what would otherwise have taken a near infinite amount of time. An example of a VSA that was briefly discussed earlier is the tensor product model[77]. We will also be discussing holographic reduced representations[62] in an upcoming section, one of the more notable and widely used VSA architectures.

\subsection{Vector components}

Components of vectors can be binary, sparse, continuous or complex. Vector operations such as addition, binding and permutation can be used to group desired elements together. Since we are looking for a representation that is specifically suited for optimizing machine learning, a number of constraints will have to be employed. With the goal to represent potentially hundreds of thousands of objects at a time, a distributed representation will have to be used, otherwise storing each component in its independent vector would not be sufficient.\\
\indent Similar objects and structures will have to be mapped to similar vectors, having a small euclidean (or Hamming) distance separating them. This makes it so that similar vectors will have a significantly greater dot product than 2 randomly chosen vectors. Such a design decision would help when filtering out the error and noise generated during the approximation and computation phases, ensuring that a word's meaning, or a pixel's color, doesn't vary too much across iterations and during recovery. This also gives the network a better ability to generalize, by finding approximate matches to novel concepts, and using its past knowledge to determine the implications of new entities.\\
\indent	Another important constraint to consider is the use of fixed length vectors. The goal here is to incorporate VSAs with standard machine learning algorithms, such as Structured Classification, where inputs and outputs are most conveniently expressed as vectors with a pre-specified number of components. Sequences, such as sentences, however don't have a fixed number of components, or a fixed structure. In the specific example of sentences, a potential solution to that problem would be using part-of-speech tagging, chunking, named entity recognition and semantic role labeling[10]. This helps us convert sentences into unambiguous data structures using the syntactic, and semantic information contained in them. Vectors' dimensions should be tuned for maximum efficiency. Larger vectors have more storage capacities, while smaller vectors decrease the system's computation time and make for faster network convergence.

\subsection{Vector operations}

All vectors being of the same size and dimension enable us to use the wide array of pre-existing vector operations and algorithms, in order to combine and structuralize objects. One such operation is the classical vector addition operator. Addition is commutative, and encodes an unordered set of vectors into a single vector of the same dimension. A structure, or sentence, could therefore be encoded in a single normalized vector, consisting of the sum of the individual word vectors. A different sentence formed using the same word vectors would also have the same resulting sum. Unlike scalar addition, vector addition preserves constituent recognition, which enables information retrieval[9]. If the dot product of the resulting sum vector with the constituent we are looking for, is close to the dimension of the vector, then that element is present in the vector. In order to better demonstrate this operation, let us consider a brief example:
}

{\fontsize{12}{18} \texttt{\color{gray!90!black} {\fontfamily{cmss}\selectfont \textls {\indent \indent Let us consider the set of n-dimensional vectors:\\
\indent \indent \indent a, b, c, d, e, and F.
	$$ F = a + b + c + d + e$$
\indent \indent \indent We would like to check if the vector $V$ is one of the vectors\\
\indent \indent  composing F.\\
\indent \indent \indent For the purpose of demonstration, let us consider:
$$V = b$$
$$ V \in F$$
\indent \indent \indent This is done by computing the dot product of the vectors\\
\indent \indent $V$ and $F:$
$$F . V = (a + b + c + d + e) * V $$
\indent \indent \indent Multiplication is distributive over addition:
\indent	$$F . V = (a + c + d + e) . V + b . V$$
\indent \indent \indent Similar vectors have a dot product that is almost equal to\\
\indent \indent their dimension $n$.
$$F . V = <mean \ 0 \ noise> + n$$}}}}

{\setlength\parindent{1cm}
\setlength\parskip{4mm}	
\noindent If the dot product of the sum vector with the tested vector is equal to, or greater, than the vector's dimension, this would indicate that the tested vector is one of the vectors that was used for the sum. However it is not possible to know exactly which vector that is, or where its position was.\\
\indent	Since vector addition stores only unordered sets of vectors, relying on it is not sufficient to encode information structure and order. We need a way to bind objects in a structure together, while retaining a certain degree of flexibility when it comes to moving them around. We want to bind words in a sentence together, while retaining the inherent meaning that comes with them if we change the order of the words in certain ways[80].

\noindent For that, we use the binding operator, to bind a sum of vectors as part of a structure description, by multiplying the sum vector by a matrix. We use the features described previously[10] to identify chunks of data that maintain coherence even when shuffled, then bind them to a matrix, or sub-matrix, depending on their relevance in the general order of things. Component recognition is done the same way in vector binding as it was using the addition operator.\\
\indent	By using multiple binding matrices, we can bind several batches, chunks, or objects into a single vector. We can then use the preserved recognition feature to determine if a certain element appears in any of them, if 2 elements appear together in the same matrix, or if they appear in the same global structure, but in different chunks. This helps subsequent learning and alleviates the ambiguity issues faced when using only vector addition.\\
\indent	In order to increase robustness, we can encode structure descriptions several times in a vector, by adding them. Structures can be represented by sequentially binding their components to matrices, in order to fix their precise order. This gives the network the ability to not only generalize to a novel structure that is similar to a the inputs, but also the ability to know how similar certain structures are, or if they were exact replicas, element and order-wise[16]. 

\subsection{Holographic Reduced Representations:}

A problem with representing concepts using connectionist networks is that items and associations are represented in different spaces. In order for part-whole hierarchies, which will be detailed later, to work as intended, we should employ a vector that acts as a reduced description of a set of vectors, and combines their characteristics into one, for easier manipulation. For this, and several other reasons stated earlier, Holographic Reduced Representations, HRR for short are used. They are constructs of the association of vectors in a reduced and compact fashion. The result of an association, of two or more vectors of a given dimension, is a vector with the same dimension[62].

\subsubsection{Matrix memory}
	
Traditional matrix memories, are simple, and have high capacities[89]. They use three basic operations for their storage needs. First off, they use encoding, where two item vectors result in a memory trace, or matrix. Secondly, they can use memory trace composition, through addition, or superposition of several memory traces, in order to compress more than two matrices, or more than two vectors, into a single matrix. Thirdly, they use a decoding process, where a memory trace and an item vector are used to give the other item vector. To better illustrate these ideas, let us showcase some vector operations.
}

{\fontsize{12}{18} \texttt{\color{gray!90!black} {\fontfamily{cmss}\selectfont \textls { \indent \indent Basic vector operations. Let us consider:\\
\\
\indent \indent $V$: Space of Vectors representing items.\\
\indent \indent $M$: Space of Matrices representing memory traces.\\
\\
\indent \indent \begin{minipage}{\textwidth}$ \boxtimes $: Encoding operation, used to transform two vectors into a  matrix:\end{minipage}
$$V \boxtimes V = M$$
\indent \indent $ \boxplus $: Trace composition operation, composes two memory traces into \indent \indent \indent \hspace{0.2mm} one:
	$$M \boxplus M = M$$
\indent \indent $\vartriangleright$: Decoding operation, used to retrieve a Vector from a Matrix, \indent \indent \indent \hspace{0.9mm} using a cue vector:
	$$V \vartriangleright M = V$$}}}}

\noindent Memory traces are very flexible. They can represent a number of associations, and compositions through pairs. Any item from any pair can be recovered, using the other vector it was associated to as cue.

{\fontsize{12}{18} \texttt{\color{gray!90!black} {\fontfamily{cmss}\selectfont \textls { 
\indent  Let a, b, c, d, e and f, be Vectors representing items.\\
\\
\indent \indent The vectors a and b can be encoded into a matrix:
	$$a \boxtimes b = M$$
\indent \indent Let us demonstrate memory trace composition:
 	$$(a \boxtimes b) \boxplus (c \boxtimes d) \boxplus (e \boxtimes f) = C$$
\indent \indent Using any vector as a decoding cue will recover the other vector of \indent \indent the pair: 
	$$e \vartriangleright C = f$$
}}}}
		
{\setlength\parindent{1cm}
\setlength\parskip{4mm}
\noindent Noise increases as more vectors are stored in a single memory trace. This leads us towards using autoassociative memory for noise and error filtering. However, the main issue with using such a representation is the exponential loss of space as dimensionality increases. To store N-dimensional vectors, $O(N^2)$ units will be needed to represent the $N \times N$ matrix. This calls for new ways, to represent vectors in memory cells using more constrict algorithms, for better space efficiency.

\subsubsection{Aperiodic convolution}

Past attempts at convolutions used aperiodic convolution for associations[77]. Two vectors, of $N$ elements each, are convoluted into a vector of $2N-1$ units. This is done by summing the trans-diagonals of the outer product of the two vectors. This vector grows with recursive convolution, as more vectors are added to it. For $n$ vectors of $N$ elements each being aperiodically convoluted, the resulting vector will be $[n*(N-1)+1]$-dimensional, which comes back to using $O(N^2)$ units, if $n$ and $N$ are equal and/or infinite. While non optimal compromises can be made, such as limiting the depth of the composition, or discarding elements outside the central ones[14], the growth problem can be avoided entirely by using circular convolution.

\subsubsection{Circular convolution}

The circular convolution[62] of two $N$-dimensional vectors, is an $N$-dimensional vector, and can be considered as a compressed outer product. The circular convolution of $n$ $N$-dimensional vectors is one $N$-dimensional vector, occupying $O(N)$ units. A representation of circular convolution is displayed in Figure \ref{fig:3}. 
\begin{wrapfigure}{r}{0.3\textwidth}
\caption {{\fontsize{9}{8}\selectfont  {\textbf {Circular Convolution} \mbox{represented} as a compressed outer product. \mbox{Image courtesy of [61].}}}}
\includegraphics[scale=0.5]{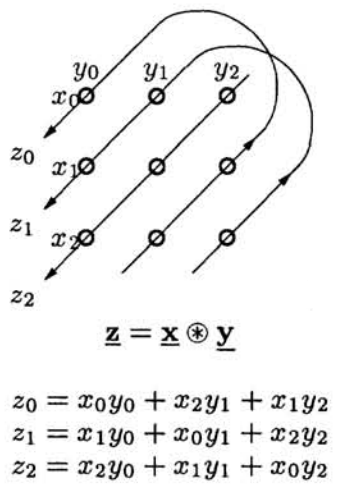}
\label{fig:3}
\end{wrapfigure} \\
Circular correlation is the approximate inverse of circular convolution, and will be able to decode a memory trace out of a vector cue, in order to give back the second item vector. However correlation only gives an approximate, noisy, version of the original item vector, which necessitates the use of autoassociative memory to filter out the noise. Circular convolution can also be approximated using Fast Fourier Transforms, for faster vector rendering.\\
\indent	In all of these methods, to successfully store a vector, we only need to store enough information to discriminate it from the other vectors. This property makes such compressions possible, up to a certain limit. The capacity of the memory model is the number of associations that can be presented usefully in a single memory trace.

\vspace{3mm}
\noindent Convolutions can also be used for variable binding, where the variable and the value are bound together, or representing sequences. Sequences can be represented in one memory trace, or divided into several memory traces depending on their lengths, and the desired accuracy[53]. The decoding process is through chaining, where the cue vector at hand will be used to start a cascade of retrievals, correlating the trace with the current item, until the end vector is reached[52]. In some cases, where elements could be shared among different structures using only one past element of the sequence may not be enough. It is therefore also possible to encode several units into the past for a more robust sequencing. To better demonstrate these concepts, let us provide an example of them:
}

{\fontsize{12}{18} \texttt{\color{gray!90!black} {\fontfamily{cmss}\selectfont \textls {\\
\\
\indent \indent $\circledast$: Circular convolution operator.
	$$a \circledast b = V$$
\indent \indent $\circledcirc$: Circular correlation operator.
	 $$b \circledcirc V = a$$
\indent \indent Sequences can be stored in several ways.\\
\\
\indent \indent Non redundant storage:
	$$X_1 \circledast a + X_2 \circledast b + X_3 \circledast c = S$$
\indent \indent Fully redundant storage, retrieval cue must be built from the ground \indent \indent up using correlations.
	$$a + a \circledast b + a \circledast b \circledast c = S$$
\indent \indent Semi-redundant storage, conserves a bit of redundancy, and gives \indent \indent much more freedom for retrieval operations.
	$$X_1 \circledast a + a \circledast b + X_2 \circledast b + b \circledast c + X_3 \circledast c = S$$
\indent \indent We can restore the memory c from S using X3 as a cue:
	 $$X_3 \circledcirc S = c$$}}}}

{\setlength\parindent{1cm}
\setlength\parskip{4mm}
\noindent Using correlation, and the relevant cue vectors, we can then retrieve our stored information from the memory network. The retrieved vectors will be noisy, so an autoassociative memory would go best with this storage method. 

\section{Working memory}

Working memory is a concept of human cognition, which aims at explaining our performance in tasks involving the manipulation of short-term information. The limitations of working memory have been thoroughly studied in psychology. Conclusive results found that memory was stored in chunks of information. The number of chunks that can be recalled at any one time was determined to be 7 $\pm$ 2 [49]. Further explorations found that the recall process depends on several other factors, including word length, syllables and prior familiarity with the words[34]. Training, however, did not seem to affect these chunks. The observed performance boosts were rather due to how chunks of information were manipulated. The biological cognitive terrain for working memory is a tripartite architecture, formed of three functionally complementary systems[4].

\subsection{Biological architecture}

The pre-frontal cortex, posterior parietal cortex, and hippocampal system are the three main components of working memory. Let us explore them briefly.\\
\indent	The pre-frontal cortex, PFC for short, specializes in the active maintenance of the internal contextual information. It is dynamically and actively updated by the basal ganglia. Basal ganglia, BG for short, are a group of subcortical gray matter, interconnected with various structures in the brain, such as the cortex, thalamus and brain stem. The PFC-BG system relies on dopamine based learning, and is able to bias on-going processing throughout the cerebral cortex. This is done in order to maintain information that is currently relevant, as well as fend off interference, and steer the attentional processes. This helps in performing task relevant processes in the face of multicentric waves of knowledge, while reducing the influence of other cortical regions which may not be relevant to the specific task at hand[48].\\
\indent	The posterior parietal cortex system performs automatic sensory and motor processing in the brain by relying on statistically accumulated knowledge. This system exhibits slow and integrative learning, through skills that are gained over long-term training. This long term processing power is believed to reside in the angular gyrus. It incorporates memory tasks involving extended periods of training and use, such as language processing, reading, writing and arithmetic operations[73].\\
\indent	The hippocampal system is a compound structure in the medial temporal lobe of the brain. It is composed of several structures, the most important of which are the hippocampus, and the entorhinal cortex. The entorhinal cortex is both the main input and output to the hippocampus. This system is specialized for rapid learning after short, or non-recurrent trials, as well as spacial localization and navigation. It binds together arbitrary information, which can be recalled by other neurological structures[42]. This structure also helps consolidate information from short-term memory, to long-term memory, through a process called Long Term Potentiation, LTP for short.

\subsection{Memory functions}

These regions support basic memory functions associated with working memory. Among the most important ones are partial recalls of on going processes, as well as controlling processing functions. They work as a central executive unit by supervising and controlling the flow of information into, and out of the system. The second of these emerges from the biasing influence by the pre-frontal cortex, which is actively maintained and updated by the basal ganglia, on the rest of the system. These two different functions are performed through the same neurological path, and distributed in a stable configuration throughout the cortex.
   
\noindent We have previously showcased two different learning mechanisms, rapid learning, and integrative learning. These operations require different types of neurons, some being rapid learners, while others being integrative statistical learners. There have been researcher that have attempted to model these neural structures, including replicating their functions and performance in basic memory recall tests such as the AX-CPT. Seeing as the PFC-BG system is the most important of these in both short lag memory and controlling other systems, we will be exploring some of its models in later sections.\\
\indent  There are several demands to creating a functional pre-frontal cortex, and basal ganglia model[23]. The system should be rapidly updating, and able to encode and maintain new information as it occurs. It must also have the ability to selectively update information, by knowing which elements to maintain in the face of interference, and which ones to update from its previous iteration. The system should have the power to condition responses in other parts of the network, by learning when to gate appropriately. As you have come to notice, these are all points that have been discussed previously in their own context, through BPTT, LSTM, NTM and PDP, but we will be seeing them combined here, in light of cognitive memory processing. 

\noindent But first, let us tackle the details of the PFC-BG operation, as well as its neurological structures and connections, in order to better understand how one goes about implementing such a system computationally.

\subsection{Memory operations in the pre-frontal cortex and basal ganglia}

The main feature of neurons in the pre-frontal cortex is their rapid update cycles. This occurs when Go neurons, in the dorsal striatum of the basal ganglia, fire. These Go units are medium spiny GABAergic neurons, which inhibit the substantia nigra pars reticulata. This leads to it releasing its tonic inhibition of the thalamus. The thalamus, being disinhibited, enables a loop of excitation into the pre-frontal cortex, which toggles the state of the bistable neurons contained within. The Go neurons are in direct competition with NoGo neurons. The NoGo neurons are also present in the dorsal striatum, and counteract the effect of Go neurons by inhibiting the external global pallidus. This excites the substantia nigra pars reticulata, which promotes the inhibition of the thalamus, therefore protecting the information stored, and contributing towards robustness in the face of interference[50]. The PFC-BG architecture is visible in Figure \ref{fig:4}. \\
	\indent The basal ganglia is connected with the pre-frontal cortex through a series of parallel loops. There are multiple separate representations, working as inner and outer loops. These loops give the PFC-BG complex the ability to process several layers of information in parallel. An outer loop will have to be maintained, serving as the framework for the inner loop's conditional statements. The outer loop is stable and has a slow rate of variance, whereas the inner loop goes through rapid iterations of changes. In order to better understand this mechanism, and its potential characteristics, let us illustrate its application through the use of a Continuous Performance Test, the AX-CPT:
}

{\fontsize{12}{18} \texttt{\color{gray!90!black} {\fontfamily{cmss}\selectfont \textls {\indent \indent  The AX-CPT task:
\\
\\
\indent \indent \indent Subjects are presented with a random sequential stimuli of \indent \indent \indent flashing letters:
	$$A or B, \textrm{followed by}  X or Y.$$
\indent \indent \indent The test sequence is of the following nature:
	$$A - Y - B - X - A - X - A - Y...etc$$
\\
\indent \indent \indent The prior stimulus should be maintained over a delay, until the \indent \indent \indent  next stimulus is presented.
\\
\\
\indent \indent \indent The target sequence is $A-X$. A dedicated button should \indent \indent \indent be pressed by the subject whenever he encounters it.
\\
\\
\indent \indent \indent When any other sequence is detected, such as: $A-Y$, $B-X$ or \indent \indent \indent $B-Y$, another button should be pressed.
\\
\\
\indent \indent \indent The first letter is therefore held in an outer loop, around \indent \indent \indent  an inner loop which houses the second letter. This is required to \indent \indent \indent detect the target sequence.	}}}}
\\
\\
\noindent These results are the fruit of several deductive studies performed on behaving animals that exhibit human-like memory performances, as well as humans. However, multiple inner loops can co-exist in one outer loop. An inner loop could also play the role of an outer loop, and house an inner loop of its own, resulting in a stack of inner loops. In order to demonstrate that, let us present a more complicated and recent version of this test, the 1-2-AX-CPT.

{\fontsize{12}{18} \texttt{\color{gray!90!black} {\fontfamily{cmss}\selectfont \textls {\indent \indent The 1-2-AX-CPT task:
\\
\\
\indent \indent \indent A $1 \ or \ 2$ stimuli is added upstream of the regular $A/B - X/Y$ \indent \indent \indent stimuli.
\\
\\
\indent \indent \indent The test sequence therefore becomes:
	$$1 - A - Y - B - X - 2 - A - X - B - Y - A - Y...etc$$
\\
\\
\indent \indent \indent \begin{minipage}{\textwidth} The target sequence here is variable:\\
\indent \indent $\bullet$ If the subject last saw a 1: $A-X$ is the target sequence.\\
\indent \indent $\bullet$ If the subject last saw a 2: $B-Y$ is the target sequence. \end{minipage}
\\
\\
\\
\indent \indent \indent The task defines an outer loop, $1 \ or \ 2$, of active \indent \indent \indent maintenance, within which resides a stack of inner loops. 
\\
\\
\indent \indent \indent Each of these inner loops is formed by an outer loop, $A \ or \ B$, \indent \indent \indent and an inner loop, $X \ or \ Y$.
}}}}
\\
\\

{\setlength\parindent{1cm}
\setlength\parskip{4mm}
\noindent A representation of the inner and outer loops is shown in Figure \ref{fig:5}. In order to prevent interference between different loops, this task is achieved through the unique architecture of the PFC. This is done through isolated, self-connected, stripes of interconnected neurons, that are capable of sustained fire, and active maintenance of the working memory. Selective updating of these stripes occurs through parallel loops that can be overwritten independently, in different areas of the basal ganglia and the pre-frontal cortex[2].

\noindent Using a dopamine-based learning mechanism, the Go and NoGo neurons are trained to appropriately gate the activity of other neurons in the striatum. Dopamine acts on these cells using different types of receptors, D1 and D2, respectively for Go and NoGo neurons. The D1-like family (D1 and D5) activate the adenyl cyclase, increasing intra-cellular cAMP levels during positive reinforcement. The D2-like family (D2, D3, and D4) inhibit the formation of cAMP during negative learning reinforcement[25].  Each neuron develops its own unique pattern of weights. cAMP in turn regulates neuronal growth and development, and mediates some behavioral responses. This dynamic gating mechanism helps switch between updating existing information relative to incoming stimulus, and protecting information in the face of interference.
\begin{figure}[t]
\begin{center}
\includegraphics[scale=0.7]{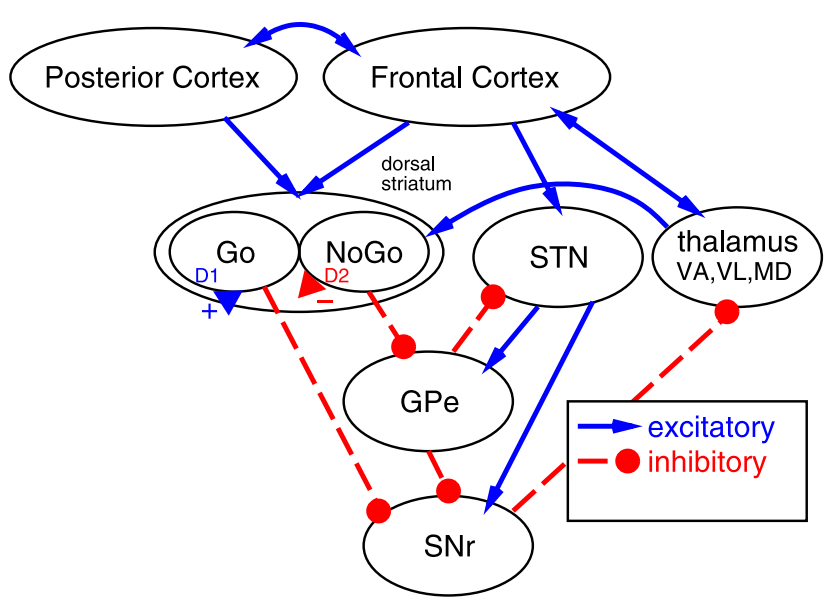}
\caption {{\fontsize{10}{8}\selectfont  {\textbf {A view on the memory architecture of the brain.} Go and NoGo represent the medium spiny neurons in the dorsal striatum. STN represents the subthalamic nucleus. The 3 segments in the thalamus are VA for ventral anterior nucleus, VL for ventral lateral nucleus, and MD for the dorso-medial nucleus. GPe is the external segment of the globus pallidus. SNr is the substantia nigra pars reticulata. The thalamus is bidirectionally excitatory with the frontal cortex. The SNr is tonically active, and inhibates the thalamus. When Go neurons activate, they inhibit the SNr, and stop the inhibition of the thalamus. NoGo neurons indirectly inhibit the SNr, by inhibiting the GPe, which usually inhibits the SNr, this in turn activates the thalamus. Image courtesy of [23].}}}
\label{fig:4}
\end{center}
\end{figure}

\subsection{Modeling the PFC-BG system}

To encode these features in a computational system, we will need a basal ganglia model, with two types of neurons, Go, and NoGo, controlling a pre-frontal cortex model. The learning problem boils down to teaching these neurons when to fire, according to the sensory input they receive from an equivalent to the posterior cortex, and the maintained pre-frontal cortex activations.\\
\indent	However, once information is encoded into the pre-frontal cortex, we would not be able to know how beneficial it was. Feedback on whether that memory was helpful or inappropriate will only come when that memory is recalled, later in time. Knowing which prior events were performed for the subsequent good, or bad performance will be hard to extrapolate, yet is critical for training our network. The pre-frontal cortex model will have to be divided into stripes, containing independent loops. The network should then decide what memories to encode in each stripe, reinforcing those who contribute to its success during memory recall. This can be done through classical neural network algorithms, such as backpropagation through time[86], although here, the problem is considerably more complex because of the different types of nodes used, the latent feedback, and the unique stripe-loop architecture.\\
\begin{figure}[t]
\begin{center}
\includegraphics[scale=1]{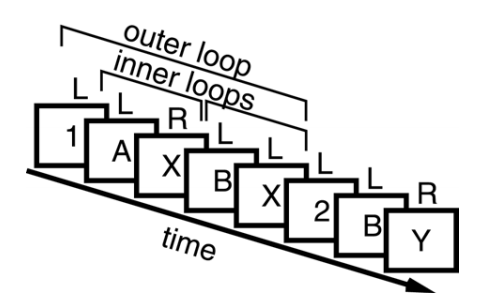}
\caption {{\fontsize{10}{8}\selectfont  {\textbf {Inner and outer loops represented through the 1-2-AX-CPT task.} The stimuli is presented in a sequence, and the participants respond by pressing one of two buttons. The right button, R, is pressed when the subject is presented with the target sequence. The left button, L, is pressed when they are presented with any other sequence. Image courtesy of [23].}}}
\label{fig:5}
\end{center}
\end{figure}
\indent	The brain solves the latent feedback problem by predicting the subsequent rewards of a certain stimulus[72]. Neurons connected to the basal ganglia, in the ventral tegmental area, and the substantia nigra pars compacta predict the dopamine stimulus, thereby reinforcing Go firing, when such maintenance usually leads to a positive reward. To model these features, we can use the "Primary Value Learned Value"[57], or PVLV, Pavlovian learning algorithm. In this algorithm, the Primary Value controls performance and learning during only the primary rewards or unconditioned stimuli. On the other hand, the learned value processes the received stimulus, and learns about the conditioned stimuli that are reliably associated with rewards.\\
\indent	The unique stripe-loop architecture can be explicitly observed in the pre-frontal cortex during working memory recall tests. This indicates that for the brain, as well as the models that should be used to model it, information that must be updated at different points in time, must also be represented in different parts of the pre-frontal cortex[38]. Anterior areas of the pre-frontal cortex are selectively activated for outer-loop information, while the dorsa-lateral parts are active for inner-loop information. The model used should therefore employ parallel loops, that are updated independently by external structures. A dynamic, context sensitive, gating mechanism should be used to update, or protect, relevant information in the face of respectively relevant, or irrelevant inputs.

\section{Modeling the cortical architecture}

\subsection{Multiple layers of representation}

In order to make the fine distinctions required to control behavior and other types of processing, the cerebral cortex is composed of 6 layers, each of which has a distinct role. The cortex needs an efficient way of adapting the synaptic weights across these multiple layers of feature-detecting neurons. BPTT is a possible model of how neural network could learn multiple layers of representations, however it is only suited to classify labeled training data, and supervised learning.\\
\indent	For this reason, a neural network that contains both bottom-up recognition connections and top down generative connections is needed. Bottom-up connections are used to recognize the input data, whereas top down generative connections fulfill their roles by generating a whole distribution of data vectors over recurrent passes. In other words, bottom-up connections help in determining the activations for certain present features, and places them in layers. Top-down connections try to generate the training data using the activations learned from the bottom-up pass. The learning algorithm used for the top-down connections is an inverted learning model using negative feedback loops[56]. These passes are done while adjusting the weightings on the connections in order to maximize performance and optimization.\\
\indent	This model is in line with several neuroscientific observations, including single cell recordings, and the theory of reciprocal connectivity between cortical areas, which indicates to a hierarchical progression of complexity in the cortex.

\subsection{Training unlabeled data}

The external data we receive as humans are random. Most of the time, there are no labeled data being presented to us. We aren't able to manually structure our brains to perform a certain task. In terms of modeling, the goal is translated as finding hidden structures in unlabeled data. This problem boils down to probabilistic density estimation. A child holds an apple, and hears it being referred to as "Apple" several times, making the connection that this object is indeed called an "Apple". A probabilistic distribution is inferred over the various possible settings of the hidden variables, by doing several recurrent passes[27]. However, a neural network with only one hidden layer is too simplistic, and would not be able to handle such rich and complex structures, nor would it be able to correctly deal with the high-dimensional operations at task. One way to solve this problem is by stacking restricted Boltzmann machines, in order to form a deep belief net. However this is not optimal. The hidden variables use discrete values, making exact inference possible, but very limited to the domain and learned scenarios, causing reduced generalization capacities.\\
	\indent Going down the probabilistic road, and simplifying our way to the lowest common denominator, our problem boils down to using either generative models, or a discriminative models. Generative models are full probabilistic models. They can be used to generate the probabilistic prediction of any, and all variables in the model. They express complex, and complete, relationships between the observed and target variable, and are able to categorize data statistically, to determine the mostly likely input that would generate such a signal. The other type of probabilistic modeling, discriminative models, only model the target variables conditionally, in views of the observed data-sets. This is a limiting probabilistic distribution, and does not give the operator insight into how the data is being generated. Its role is to simply categorize a signal.\\
	\indent For a long time, the general consensus was that generative models were inferior to their discriminative counterparts[36], however that turned out to be false. Generative models, while they may take more time to resolve, have superior computation powers, due to their transparency towards the full probabilistic distribution. A generative model can infer a discriminative model from its data, whereas the opposite is not true.
	
\noindent Modeling the mind should therefore rely on unsupervised learning, using many hidden layers in a non-linear distributive neural network. Networks build a progressively more complex hierarchy using sensory data, by passing it through a generative model. In order to solve the problem of reciprocal connectivity through top-down and bottom-up connections, a good candidate to that are PCSPs. We will talk briefly about them, before moving onto a larger concept which makes use of them.

\subsection{Parallel constraint satisfaction processes}

Parallel constraint satisfaction processes[65], PCSP for short, are a type of network used to model cognitive dissonance, and other contemporary issues in social, and Gestalt psychology. Activations pass around symmetrically connected nodes, until the activation of all the nodes relaxes into a state. What we mean by all nodes “relax” is that all constraints are met when the change percentage between the previous, and current state of each node asymptotes. This final state satisfies all the constraints set among the nodes, which are connected with each otehrs, and represent hypotheses detailing the presence or absence of certain features. These connections serve as a ground to prove hypotheses as being either consistent within each others, or inconsistent and contradictory.\\ 
\indent Activations spread in parallel among the nodes. Nodes with positive links, or connections, will activate each others reciprocally. Nodes with negative links inhibit each others. This results in a global and viable solution to the constraints among the entire set of nodes. 

\subsection{Part-whole hierarchies}

Representing simple structures in a distributive representation network is complex. Objects are not symbols, or singular entities, they are instead represented as patterns across many units. It is even harder to represent complex structures, such as sentences, where the meaning is composed of several constituents, with relations interweaving across several of it sub-elements. Attentional focus must therefore be brought on the constituents of that structure, while at the same time maintaining the whole meaning in check. To deal with that problem, one potential solution is using part-whole hierarchies, where objects at one level are composed of inter-related objects at the next level down.\\
\indent	In the classical digital computer, hierarchical data structures are composed of a set of fields, or variables, containing pointers to the content of the fields, their values. Addresses are a small representation of the symbol they point to, and many symbols can be put together to create a fully-articulated representation. This simple representation comes at the cost of being limited by the Von Neumann bottleneck[5], in which the system runs out of hardware space.\\
\indent	A famous account of disagreement between connectionists and symbolic architecture advocates is that of Fodor and Pylyshyn[15]. They believed that the mind operates around rule governed formulations, and the manipulation of sentences in an inner linguistic code. Connectionism was seen as a step backwards, devoid of any real computation, resembling associationism, where an input is associated with its output through trial and error. However, as more progress was being made in both human cognition, and connectionism, many of their points didn't hold through. Neural networks can generalize to non test-case scenarios, create an internal representation to express regularities in the domain, and use isomorphisms in order to encode structures more efficiently. This indicates that distributed representations have much more power than previously thought.\\
\indent	In a PDP network, patterns also allow remote access to fuller representations. Most processing can be done using parallel constraint satisfaction processes, in which an activation passes around symmetrically connected nodes, until the states satisfy the constraints set among the nodes. 

\noindent PCSP results in an inflexible inner loop, leading to two different ways of performing inference[65].

\subsubsection{Intuitive inferences, and rational inferences}

Inferences are performed efficiently by allowing the neural network to settle down into a stable state. This allows the network to reach a conclusive result. Inferences are influenced by the network's weightings, past knowledge, and connection strength. Some operations require a single settling of the network to resolve, such cases are referred to as intuitive inferences.\\
\indent	Other operations require a more serial approach, where several intuitive inferences are performed in sequence, this is known as rational inferences. Rational inferences involve sequences of changes in the way in which parts of the task are mapped into the network. It is important to note that a single task can sometimes be performed in different ways, either requiring intuitive inference, or rational inference, depending on the degree of complexity of its iterations.

\subsubsection{Internal network structure}

In order to build a part-whole hierarchy, it is possible to map out the internal representation of the network by hand. The algorithm then sorts the inputs, applies the necessary steps, and derives conclusions using our base framework. It is also possible to give the network the freedom to use its experience of a set of propositions to construct its own internal representation of concepts. Propositions are presented in a neutral way, and the network translates the input into active units, using its own associations. Similar patterns of activity will therefore represent similar relations, and the representation will be built around correlations between the given input. This leads to better generalization due to the network searching for features that make it simple to express the regularities of the domain. \\
\indent	Instead of using a separate bit for each possible object that could be stored in the memory, each bit is shared between many objects. A bit is applied to one position of the object, the combination of its identity and position within the object activates a role specific unit that contributes to the recognition of the object. The recognition model is shared across the entities, within one level. This model is called within-level timesharing[45].\\
\indent	In order to share knowledge in the connections between different levels in the part-whole hierarchy, it is necessary to implement flexible mapping of pieces of the task into a module of the network. This is done by choosing a node as the current whole, the other units are devoted to either describing its global properties, or describing its major constituents. The pattern of activity that represents a same object is different depending on the whole, or context, it was presented in[28]. This implementation readily violates the assumption that each entity in the world has a unique corresponding representation in the network, which was an essential part of classical symbolic architectures. One system that hits a lot of these points, and provides a good biologically plausible architecture for computing is HTMs.

\subsection{Hierarchical Temporal Memory}

Hierarchical Temporal Memory[22], HTM for short, is a bio-inspired machine learning model, that makes use of distributed representations to store information. Its main goal is modeling the neocortex, using nodes that are arranged in columns, layers, regions, and in a hierarchy. 
	
\noindent It is compromised of two separate stages, a spatial pooler, and a temporal pooler. The spatial pooler creates a sparse binary representation from the network's input. The temporal pooler makes predictions, that are not provided in the input, in response to the vector sequences. The predictions can range from next step predictions, to predictions into the future several steps ahead.\\
\indent	The model represents dendrites using non-linear segments, and mimics them using a sum and threshold function. These structures connect with other neurons, or nodes, forming an equivalent of a synapse, leading a node to fire when enough activations have been received by any segments, through an OR operator. The segments, and their weights can be learned, and changed adaptively with the received inputs. Their number for a given node, as well as their total number in the structure can change with time to accommodate machine learning.\\
\indent	HTMs are trained on large sets of patterns and sequences, where information is stored in a distributed fashion, following what was described previously. The network has the freedom to choose how, and where the information is stored, making for better generalization. HTMs have VSA features[58], and are able to achieve bundling, by storing several objects in one vector. They are also able to achieve variable binding, where a role can be used to obtain the values of the filler.

\section{Constructing the Code: A recap}

Most of the work that has been carried out in the field of artificial neural networks, and presented here, has targeted very specific features. From language processing, to visual imagery, and models of memory, none of these models encompass the entirety of the mind. Using the limited knowledge presented in this review, we will attempt to make our own model of a general brain architecture. As we have previously seen, the brain is limited by several factors, working memory is one example of them. In the future we may want to exceed them, but for now, to insure a one to one scale, we will work within these boundaries.\\
\indent	Symbolic and localist architectures are not flexible enough, and would not be able to deal with the large amounts of data that our brains are presented with on a daily basis. For this reason, our model will be using a distributed representation. Entities will be represented by patterns, and spread across many nodes. Variable binding is an essential part of human cognition, but it used to be a challenge to deal with in distributed connectionist models. However vector symbolic architectures have proved that they can handle variable binding to a fairly decent degree. In order to encode enough features, and have enough robustness and plasticity in our networks, we will opt to use high-dimensional vectors. The high-dimensional vectors used should all be of the same dimension, to enable us to use the wide array of pre-existing, and thoroughly tested mathematical knowledge on them. Our brains are interconnected, with constant signal propagation. We will therefore model them using a recurrent system of artificial neural networks. 

\noindent Although there is a large number of neuron types in the brain, most of them are due to their anatomical, morphological and chemical differences. The two types of neurons that absolutely have to be accounted for, are the classical specificity neurons, and mixed selectivity neurons. Classical specificity neurons are simple, and integrate activity coming from different regions in a linear way, in order to activate, or inhibit a certain structure. They respond specifically to one stimulus or task. Unfortunately, currently, there is not enough data to determine exactly where the classical neurons reside in the brain, and exactly what are their advantages over mixed selectivity neurons. For this purpose, and the purpose of simplification, we will assume that classical neurons are used exclusively for the input, and output of data. They will be modeled using linear nodes, and will help transfer data into, and out of the neural architecture.\\
\indent	Mixed selectivity neurons are overwhelmingly present in higher-order structures. They have complex, and non-linear activation patterns. These are the neurons responsible for most of the logical computation done in our brain. Therefore, they will be modeled as hidden nodes, with non-linear activation functions. Our input and output layers will be classical selectivity neurons, represented by linear nodes. Our hidden layers will be using mixed selectivity neurons, represented by non-linear nodes. Nodes will use a sum and threshold function, through a leaky integrate and fire model. They will be connected with each others by links, playing the role of synapses. The number of links will be variable, and subject to machine learning. New connections can be created or dismantled after long-term training.  

\noindent Inserting gating nodes seems an essential part of any neural network. We do not want oscillating weights. We do not want any gradients to vanish, or explode. And most importantly, we do not want the network to be heavily influenced by recent inputs and modifications. However, we doubt there are gating neurons in the brain. As a result, all nodes, whether they are gates, memory cells, or regular neurons, will have to be identical. It is therefore necessary to use flexible mappings between the entities. Each node will play the role of the whole, and all the units connecting to it will be used to gate information in, and out of it. Each node will have its own set of representations, and a representation of a certain object will be different between two nodes. Using these characteristics, it is theoretically possible to have each node behave differently in response to the signal it is receiving. A node will know if it is being used as a gate node, or computing node, or both at the same time in response to different clusters of nodes, based on the input pattern it has received, and based on its previous activations. However, how this will be achieved mathematically remains to be seen.\\
\indent	The cortex is divided into separate areas, and the concept that each area controls a more or less specific functions has been tested several time, and is agreed upon. In all humans, the visual cortex is located in the Brodmann areas 17, 18, 19, while the auditory cortex is located in Brodmann areas 41 and 42. However we believe that this is not due to a special internal structure of the brain. This is simply due to how exterior stimuli is fed into the brain following its specific paths. Although not discussed here, these areas can change if the signal was rerouted from one structure to others[59]. 

\noindent But one characteristic that is constant, for a seemingly unwarranted reason, is the distinction between the cortex, and the central gray nuclei(CGNs). CGNs are an interconnected set of neurons that compile activations from different areas of the brain, perform their computations, then reroute them to other structures. It is hard to believe that such unique, yet consistent elements would arise through training. We are led to believe that there is an internal predisposition to such a structure, and we should therefore enforce it into our model.\\
\indent	The cerebral cortex has a laminar structure, and is composed of 6 layers forming microcircuits. These microcircuits are grouped into cortical minicolumns. Each layer has special characteristic connections with the others. This can be seen rather constantly through the cortex. For this purpose, we will be stacking unrestricted Boltzmann machines to model this architecture. Unrestricted Boltzmann Machines are identical to RBMs, with the exception of the presence of connections between hidden units. Central grey nuclei will be modeled using fully interconnected, or highly interconnected recurrent neural networks. The connections should be flexible and differentiable end to end. The network will be trained through unsupervised learning, using unlabeled data, in ways comparable to how the brain would receive its data. Because there are generally no restrictions on the network, the network will construct its own internal representation. Similar entities will end up with similar vectors.\\ 
\indent	In order to effectively deal with the extremely large amounts of data that a model such as the brain is bombarded with on a daily basis, we will be using Holographic Reduced Representations to conserve memory space. A variable, and its value will be bound together through circular convolution, in order to form one vector. Vectors can be compressed together to sequence them, or indicate similar entities. This will also allow us to maintain the sequential recall order for complex tasks. In order to save time, and be able to increment steps at a higher rate, we will be using approximate operations. This includes using FFT instead of circular convolution on HRRs. These approximations will require a clean-up memory, so an autoassociative storage method will be employed.

\noindent As we have seen throughout this review, tremendous ground has been covered, and astonishing work has been done so far in the field of computational neuroscience. How far we will be able to go in the following years remains to be seen. The path to the future is still very much obscure, but with the recent surge of interest in the concept of mind modeling, and the combined efforts of researchers from several subspecialities, this goal will definitely be within our reach in the near future.
}
\newpage
\subsubsection*{References}

\noindent [1]: Ackley, David H., Geoffrey E. Hinton, and Terrence J. Sejnowski. "A learning algorithm for boltzmann machines." Cognitive science 9.1 (1985): 147-169.

\noindent [2]: Alexander, Garrett E., Mahlon R. DeLong, and Peter L. Strick. "Parallel organization of functionally segregated circuits linking basal ganglia and cortex." Annual review of neuroscience 9.1 (1986): 357-381.

\noindent [3]: Asaad, Wael F., Gregor Rainer, and Earl K. Miller. "Neural activity in the primate prefrontal cortex during associative learning." Neuron 21.6 (1998): 1399-1407.

\noindent [4]: Atallah, Hisham E., Michael J. Frank, and Randall C. O'Reilly. "Hippocampus, cortex, and basal ganglia: Insights from computational models of complementary learning systems." Neurobiology of learning and memory 82.3 (2004): 253-267.

\noindent [5]: Bacus, J. "Can programming be liberated from the von Neuman style." Comm. ACM 21 (1978): 899. 

\noindent [6]: Bellman, Richard, et al. Adaptive control processes: a guided tour. Vol. 4. Princeton: Princeton university press, 1961. 

\noindent [7]: Bengio, Yoshua, Patrice Simard, and Paolo Frasconi. "Learning long-term dependencies with gradient descent is difficult." Neural Networks, IEEE Transactions on 5.2 (1994): 157-166.

\noindent [8]: Bengio, Yoshua, and Samy Bengio. "Modeling High-Dimensional Discrete Data with Multi-Layer Neural Networks." NIPS. Vol. 99. 1999.

\noindent [9]: Caid, William R., Susan T. Dumais, and Stephen I. Gallant. "Learned vector-space models for document retrieval." Information processing and management31.3 (1995): 419-429. 

\noindent [10]: Collobert, Ronan, et al. "Natural language processing (almost) from scratch."The Journal of Machine Learning Research 12 (2011): 2493-2537. 

\noindent [11]: Das, Sreerupa, C. Lee Giles, and Guo-Zheng Sun. "Learning context-free grammars: Capabilities and limitations of a recurrent neural network with an external stack memory." Proceedings of The Fourteenth Annual Conference of Cognitive Science Society. Indiana University. 1992. 

\noindent [12]: Desjardins, Guillaume, et al. "Tempered Markov chain Monte Carlo for training of restricted Boltzmann machines." International Conference on Artificial Intelligence and Statistics. 2010. 

\noindent [13]: Donoho, David L. "High-dimensional data analysis: The curses and blessings of dimensionality." AMS Math Challenges Lecture (2000): 1-32. 

\noindent [14]: Eich, Janet M. "A composite holographic associative recall model."Psychological Review 89.6 (1982): 627. 

\noindent [15]: Fodor Jerry A., and Zenon W. Pylyshyn. "Connectionism and cognitive architecture: A critical analysis." Cognition 28.1 (1988): 3-71. 

\noindent [16]: Gallant, Stephen I., and T. Wendy Okaywe. "Representing objects, relations, and sequences." Neural computation 25.8 (2013): 2038-2078. 

\noindent [17]: Gers, Felix A., Jürgen Schmidhuber, and Fred Cummins. "Learning to forget: Continual prediction with LSTM." Neural computation 12.10 (2000): 2451-2471. 

\noindent [18]: Gers, Felix, and Jürgen Schmidhuber. "Recurrent nets that time and count."Neural Networks, 2000. IJCNN 2000, Proceedings of the IEEE-INNS-ENNS International Joint Conference on. Vol. 3. IEEE, 2000. 

\noindent [19]: Giles, C. Lee, et al. "Learning and extracting finite state automata with second-order recurrent neural networks." Neural Computation 4.3 (1992): 393-405. 

\noindent [20]: Graves Alex, Greg Wayne, and Ivo Danihelka. "Neural Turing Machines." arXiv preprint arXiv:1410.5401 (2014).

\noindent [21]: Hadley, Robert F. "The problem of rapid variable creation." Neural computation21.2 (2009): 510-532.

\noindent [22]: Hawkins, Jeff, Subutai Ahmad, and Donna Dubinsky. "Hierarchical temporal memory including HTM cortical learning algorithms." Techical report, Numenta, Inc, Palto Alto(2010). 

\noindent [23]: Hazy, Thomas E., Michael J. Frank, and Randall C. O’Reilly. "Banishing the homunculus: making working memory work." Neuroscience 139.1 (2006): 105-118.

\noindent [24]: Hebb, Donald Olding. The organization of behavior: A neuropsychological theory. Psychology Press, 2005. 

\noindent [25]: Hernández-López, Salvador, et al. "D1 receptor activation enhances evoked discharge in neostriatal medium spiny neurons by modulating an L-type Ca2+ conductance." The Journal of neuroscience 17.9 (1997): 3334-3342.

\noindent [26]: Hinton, Geoffrey E. "Learning distributed representations of concepts."Proceedings of the eighth annual conference of the cognitive science society. Vol. 1. 1986. 

\noindent [27]: Hinton, Geoffrey E. "Learning multiple layers of representation." Trends in cognitive sciences 11.10 (2007): 428-434. 

\noindent [28]: Hinton, Geoffrey E. "Mapping part-whole hierarchies into connectionist networks." Artificial Intelligence 46.1 (1990): 47-75.

\noindent [29]: Hochreiter, Sepp, and Jürgen Schmidhuber. "Long short-term memory." Neural computation 9.8 (1997): 1735-1780. 

\noindent [30]: Hodgkin, Alan L., and Andrew F. Huxley. "A quantitative description of membrane current and its application to conduction and excitation in nerve."The Journal of physiology 117.4 (1952): 500-544. 

\noindent [31]: Holzer, Adrian, Patrick Eugster, and Benoît Garbinato. "Evaluating implementation strategies for location-based multicast addressing." Mobile Computing, IEEE Transactions on 12.5 (2013): 855-867.

\noindent [32]: Hopfield, John J. "Neural networks and physical systems with emergent collective computational abilities." Proceedings of the national academy of sciences 79.8 (1982): 2554-2558. 

\noindent [33]: Hsu, Yuan-Yih, and Chien-Chuen Yang. "A hybrid artificial neural network-dynamic programming approach for feeder capacitor scheduling." Power Systems, IEEE Transactions on 9.2 (1994): 1069-1075.

\noindent [34]: Hulme, Charles, et al. "The role of long-term memory mechanisms in memory span." British Journal of Psychology 86 (1995): 527. 

\noindent [35]: Ingster, Yuri I., Christophe Pouet, and Alexandre B. Tsybakov. "Classification of sparse high-dimensional vectors." Philosophical Transactions of the Royal Society of London A: Mathematical, Physical and Engineering Sciences367.1906 (2009): 4427-4448. 

\noindent [36]: Jordan, A. "On discriminative vs. generative classifiers: A comparison of logistic regression and naive bayes." Advances in neural information processing systems 14 (2002): 841. 

\noindent [37]: Kanerva, Pentti. "Hyperdimensional computing: An introduction to computing in distributed representation with high-dimensional random vectors." Cognitive Computation 1.2 (2009): 139-159. 

\noindent [38]: Koechlin, Etienne, et al. "Dissociating the role of the medial and lateral anterior prefrontal cortex in human planning." Proceedings of the National Academy of Sciences 97.13 (2000): 7651-7656. 

\noindent [39]: Kohonen, Teuvo. "Correlation matrix memories." Computers, IEEE Transactions on 100.4 (1972): 353-359. 

\noindent [40]: Kosko, Bart. "Bidirectional associative memories." Systems, Man and Cybernetics, IEEE Transactions on 18.1 (1988): 49-60. 

\noindent [41]: Levitt, Jonathan B., et al. "Topography of pyramidal neuron intrinsic connections in macaque monkey prefrontal cortex (areas 9 and 46)." Journal of Comparative Neurology 338.3 (1993): 360-376. 

\noindent [42]: Maass, Anne, et al. "Laminar activity in the hippocampus and entorhinal cortex related to novelty and episodic encoding." Nature communications 5 (2014). 

\noindent [43]: Malsburg, Cvd. "The correlation theory of brain function." Internal Report MPI für biophysikalische Chemie 81 (1981). 

\noindent [44]: Marcus, G. F. "The algebraic mind: Reﬂections on connectionism and cognitive science." (2000).  Cambridge, MA: MIT Press. 

\noindent [45]: McClelland, James L., and David E. Rumelhart. "An interactive activation model of context effects in letter perception: I. An account of basic findings."Psychological review 88.5 (1981): 375. 

\noindent [46]: McClelland, James L., David E. Rumelhart, and Geoffrey E. Hinton. "The appeal of parallel distributed processing." Cambridge, MA: MIT Press, 1986. 

\noindent [47]: McCulloch, Warren S., and Walter Pitts. "A logical calculus of the ideas immanent in nervous activity." The bulletin of mathematical biophysics 5.4 (1943): 115-133. 

\noindent [48]: McNab Fiona, and Torkel Klingberg. "Prefrontal cortex and basal ganglia control access to working memory." Nature neuroscience 11.1 (2008): 103-107. 

\noindent [49]: Miller, George A. "The magical number seven, plus or minus two: some limits on our capacity for processing information." Psychological review 63.2 (1956): 81. 

\noindent [50]: Mink, Jonathan W. "The basal ganglia: focused selection and inhibition of competing motor programs." Progress in neurobiology 50.4 (1996): 381-425. 

\noindent [51]: Monner, Derek, and James A. Reggia. "A generalized LSTM-like training algorithm for second-order recurrent neural networks." Neural Networks 25 (2012): 70-83. 

\noindent [52]: Murdock, Bennet B. "A distributed memory model for serial-order information."Psychological Review 90.4 (1983): 316. 

\noindent [53]: Murdock, Bennet B. "Serial-order effects in a distributed-memory model." (1987).  In David S. Gorfein and Robert R. Ho man, editors, memory and learninig. The Ebbinghaus Centennial Conference, pages 277{310. Lawrence Erlbaum Associates 

\noindent [54]: Neumann, Jane. "Learning the systematic transformation of holographic reduced representations." Cognitive Systems Research 3.2 (2002): 227-235. 

\noindent [55]: Newell, Allen. "Physical Symbol Systems*." Cognitive science 4.2 (1980): 135-183. 

\noindent [56]: Oh Jong-Hoon, and H. Sebastian Seung. "Learning Generative Models with the Up Propagation Algorithm." Advances in Neural Information Processing Systems. 1998. 

\noindent [57]: O'Reilly, Randall C., et al. "PVLV: the primary value and learned value Pavlovian learning algorithm." Behavioral neuroscience 121.1 (2007): 31. 

\noindent [58]: Padilla, Daniel E., and Mark D. McDonnell. "A neurobiologically plausible vector symbolic architecture." Semantic Computing (ICSC), 2014 IEEE International Conference on. IEEE, 2014. 

\noindent [59]: Pallas, Sarah L., Anna W. Roe, and Mriganka Sur. "Visual projections induced into the auditory pathway of ferrets. I. Novel inputs to primary auditory cortex (AI) from the LP/pulvinar complex and the topography of the MGN‐AI projection." Journal of Comparative Neurology 298.1 (1990): 50-68. 

\noindent [60]: Pineda, Fernando J. "Dynamics and architecture for neural computation."Journal of Complexity 4.3 (1988): 216-245. 

\noindent [61]: Plate, Tony "Estimating analogical similarity by dot-products of Holographic Reduced Representations." Advances in neural information processing systems (1994): 1109-1109. 

\noindent [62]: Plate, Tony. "Holographic Reduced Representations: Convolution Algebra for Compositional Distributed Representations." IJCAI. 1991. 

\noindent [63]: Pollack, Jordan B. "On connectionist models of natural language processing." Computing Research Laboratory, New Mexico State University, 1987. 

\noindent [64]: Pollack, Jordan B. "Recursive distributed representations." Artificial Intelligence46.1 (1990): 77-105. 

\noindent [65]: Read Stephen J., Eric J. Vanman, and Lynn C. Miller. "Connectionism, parallel constraint satisfaction processes, and gestalt principles:(Re) introducing cognitive dynamics to social psychology." Personality and Social Psychology Review 1.1 (1997): 26-53. 

\noindent [66]: Rigotti, Mattia, et al. "Internal representation of task rules by recurrent dynamics: the importance of the diversity of neural responses." Frontiers in computational neuroscience 4 (2010). 

\noindent [67]: Rigotti, Mattia, et al. "The importance of mixed selectivity in complex cognitive tasks." Nature 497.7451 (2013): 585-590.  

\noindent [68]: Robinson, A. J., and Frank Fallside. The utility driven dynamic error propagation network. University of Cambridge Department of Engineering, 1987. 

\noindent [69]: Rosenfeld, Ronald, and David S. Touretzky. "Coarse-coded symbol memories and their properties." Complex Systems 2.4 (1988): 463-484. 

\noindent [70]: Roweis, Sam. "Boltzmann machines." lecture notes (1995).

\noindent [71]: Rudolph-Lilith, Michelle, and Lyle E. Muller. "Aspects of randomness in biological neural graph structures." BMC Neuroscience 14.Suppl 1 (2013): P284. 

\noindent [72]: Schultz, Wolfram. "Predictive reward signal of dopamine neurons." Journal of neurophysiology 80.1 (1998): 1-27. 

\noindent [73]: Seghier, Mohamed L. "The angular gyrus multiple functions and multiple subdivisions." The Neuroscientist 19.1 (2013): 43-61.  

\noindent [74]: Siegelmann, Hava T., and Eduardo D. Sontag. "On the computational power of neural nets." Journal of computer and system sciences 50.1 (1995): 132-150. 

\noindent [75]: Smolensky, Paul. "On the proper treatment of connectionism." Behavioral and brain sciences 11.01 (1988): 1-23.

\noindent [76]: Smolensky, Paul. "Parallel distributed processing: explorations in the microstructure of cognition, vol. 1. chapter Information processing in dynamical systems: foundations of harmony theory." MIT Press, Cambridge, MA, USA 15 (1986): 18. 

\noindent [77]: Smolensky, Paul. "Tensor product variable binding and the representation of symbolic structures in connectionist systems." Artificial intelligence 46.1 (1990): 159-216. 

\noindent [78]: Sowell, Elizabeth R., et al. "Mapping cortical change across the human life span." Nature neuroscience 6.3 (2003): 309-315. 

\noindent [79]: Touretzky, David S., and Geoffrey E. Hinton. "A distributed connectionist production system." Cognitive Science 12.3 (1988): 423-466. 

\noindent [80]: Treisman, Anne. "Solutions to the binding problem: progress through controversy and convergence." Neuron 24.1 (1999): 105-125. 

\noindent [81]: Turing, Alan Mathison. "On computable numbers, with an application to the Entscheidungsproblem." J. of Math 58.345-363 (1936): 5.

\noindent [82]: Turing, Alan Mathison. "On computable numbers, with an application to the Entscheidungsproblem. A correction." Proceedings of the London Mathematical Society 2.1 (1938): 544-546.

\noindent [83]: Verleysen, Michel. "Learning high-dimensional data." Nato Science Series Sub Series III Computer And Systems Sciences 186 (2003): 141-162.

\noindent [84]: Von Neumann, John. "First Draft of a Report on the EDVAC." IEEE Annals of the History of Computing 4 (1993): 27-75. 

\noindent [85]: Wagemans, Johan, et al. "A century of Gestalt psychology in visual perception: II. Conceptual and theoretical foundations." Psychological bulletin 138.6 (2012): 1218.  

\noindent [86]: Werbos, Paul J. "Generalization of backpropagation with application to a recurrent gas market model." Neural Networks 1.4 (1988): 339-356. 

\noindent [87]: Weston, Jason, Sumit Chopra, and Antoine Bordes. "Memory networks." arXiv preprint arXiv:1410.3916 (2014).  

\noindent [88]: Williams, Ronald J., and David Zipser. "A learning algorithm for continually running fully recurrent neural networks." Neural computation 1.2 (1989): 270-280. 

\noindent [89]: Willshaw, David. "Holography, associative memory, and inductive generalization." (1985). 

\noindent [90]: Wolpert, D. "A computationally universal field computer wich is purely linear." Relatório Técnico. LA-UR-91-2937, Los Alamos National Laboratory, 1991. 

\noindent [91]: Yarom, Yosef, and Jorn Hounsgaard. "Voltage fluctuations in neurons: signal or noise?." Physiological Reviews 91.3 (2011): 917-929.

\end{document}